\documentclass[twocolumn, switch]{article} 

\usepackage{preprint}

\usepackage{amsmath, amsthm, amssymb, amsfonts}

\usepackage[numbers,square]{natbib}
\usepackage[utf8]{inputenc}	
\usepackage[T1]{fontenc}	
\usepackage{xcolor}		
\usepackage[colorlinks = True,
            linkcolor = purple,
            urlcolor  = black,
            citecolor = black,
            anchorcolor = black,
            allcolors = black]{hyperref}	
\usepackage{booktabs} 		
\usepackage{nicefrac}		
\usepackage{microtype}		
\usepackage{float}			

\usepackage{enumitem}
\usepackage{subcaption}
\usepackage{multirow}

\newcommand{\itab}[1]{\hspace{0em}\rlap{#1}}
\newcommand{\tab}[1]{\hspace{.13\textwidth}\rlap{#1}}
\newcommand{\rulesep}{\unskip\ \vrule\ }

\usepackage{array}
\newcolumntype{C}[1]{>{\centering\arraybackslash}p{#1}}

\usepackage{titlesec}
\titlespacing\section{0pt}{12pt plus 3pt minus 3pt}{1pt plus 1pt minus 1pt}
\titlespacing\subsection{0pt}{10pt plus 3pt minus 3pt}{1pt plus 1pt minus 1pt}
\titlespacing\subsubsection{0pt}{8pt plus 3pt minus 3pt}{1pt plus 1pt minus 1pt}

\usepackage{tikz,xcolor,hyperref}

\definecolor{lime}{HTML}{A6CE39}
\DeclareRobustCommand{\orcidicon}{
	\begin{tikzpicture}
	\draw[lime, fill=lime] (0,0)
	circle [radius=0.16]
	node[white] {{\fontfamily{qag}\selectfont \tiny ID}};
	\draw[white, fill=white] (-0.0625,0.095)
	circle [radius=0.007];
	\end{tikzpicture}
	\hspace{-2mm}
}
\foreach \x in {A, ..., Z}{\expandafter\xdef\csname orcid\x\endcsname{\noexpand\href{https://orcid.org/\csname orcidauthor\x\endcsname}
			{\noexpand\orcidicon}}
}

\title{Learning-Based Autonomous Navigation, Benchmark Environments and Simulation Framework for Endovascular Interventions}

\usepackage{authblk}

\author[1]{Lennart~Karstensen\orcidA{}}
\author[2]{Harry~Robertshaw\orcidB{}}
\author[3]{Johannes~Hatzl\orcidC{}}
\author[2]{Benjamin~Jackson}
\author[4]{Jens~Langejürgen\orcidD{}}
\author[1,5]{Katharina~Breininger\orcidE{}}
\author[6]{Christian~Uhl}
\author[2]{S.M.Hadi~Sadati}
\author[2,7]{Thomas~Booth}
\author[2]{Christos~Bergeles\orcidF{}}
\author[1]{Franziska~Mathis-Ullrich\orcidG{}}

\affil[1]{Department Artificial Intelligence in Biomedical Engineering, Friedrich-Alexander-University, Erlangen, 91052, Germany, \{franziska.mathis-ullrich@fau.de\}}
\affil[2]{School of Biomedical Engineering and Imaging Sciences, King’s College London, London, WC2R 2LS,  United Kingdom}
\affil[3]{Department of Vascular and Endovascular Surgery, University Hospital Heidelberg, Heidelberg, 69120,  Germany}
\affil[4]{Clinical Health Technologies, Fraunhofer IPA, Mannheim, 68167,  Germany}
\affil[5]{Center for AI and Data Science, Julius-Maximilians-Universität Würzburg, Würzburg, 97074, Germany}
\affil[6]{Department of Vascular Surgery, University Hospital RWTH Aachen, Aachen, 52074,  Germany}
\affil[7]{Department of Neuroradiology, Kings College Hospital, Ruskin Wing, London, SE5 9RS, United Kingdom}

\begin{document}

\twocolumn[ 
  \begin{@twocolumnfalse} 

\maketitle

\begin{abstract}
Endovascular interventions are a life-saving treatment for many diseases, yet they suffer from drawbacks such as radiation exposure and potential scarcity of proficient physicians.
Robotic assistance during these interventions could be a promising support towards these problems. 
Research focusing on autonomous endovascular interventions utilizing artificial intelligence-based methodologies is gaining popularity. 
However, variability in assessment environments hinders the ability to compare and contrast the efficacy of different approaches, primarily due to each study employing a unique evaluation framework. 
In this study, we present deep reinforcement learning-based autonomous endovascular device navigation on three distinct digital benchmark interventions: BasicWireNav, ArchVariety, and DualDeviceNav. 
The benchmark interventions were implemented with our modular simulation framework stEVE (simulated EndoVascular Environment).
Autonomous controllers were trained solely in simulation and evaluated in simulation and on physical test benches with camera and fluoroscopy feedback.
Autonomous control for BasicWireNav and ArchVariety reached high success rates and was successfully transferred from the simulated training environment to the physical test benches, while autonomous control for DualDeviceNav reached a moderate success rate.
The experiments demonstrate the feasibility of stEVE and its potential for transferring controllers trained in simulation to real-world scenarios. 
Nevertheless, they also reveal areas that offer opportunities for future research. 
This study demonstrates the transferability of autonomous controllers from simulation to the real world in endovascular navigation and lowers the entry barriers and increases the comparability of research on endovascular assistance systems by providing open-source \href{https://github.com/lkarstensen/stEVE_training}{training scripts}, \href{https://github.com/lkarstensen/stEVE_bench}{benchmarks} and the \href{https://github.com/lkarstensen/stEVE}{stEVE framework}.

\end{abstract}
\vspace{0.35cm}

  \end{@twocolumnfalse} 
] 



\section{Introduction}

\begin{figure*}[t]
    \centering
    \includegraphics[width=\hsize]{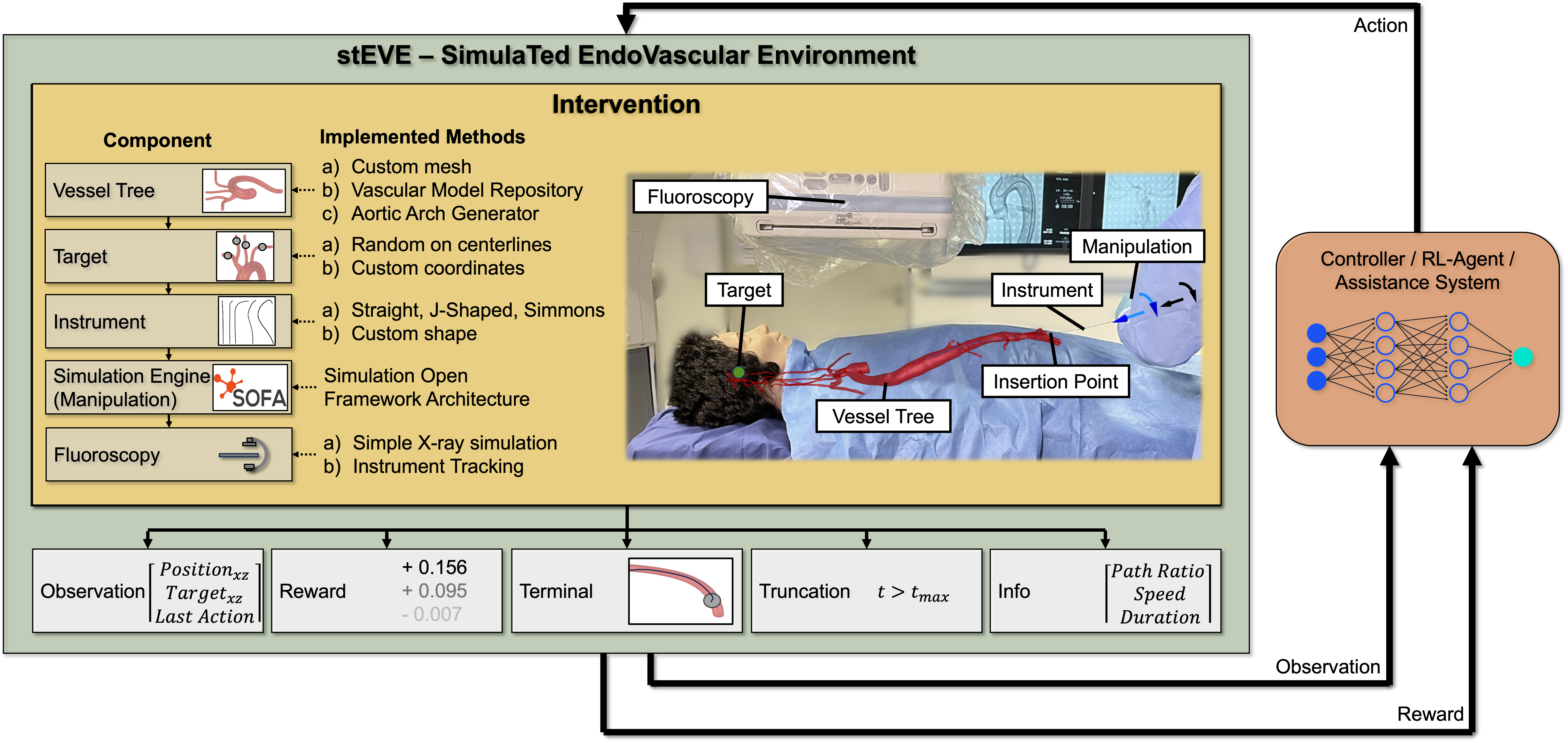}
    \caption{Structure of stEVE and its components derived from the illustrated endovascular intervention, the reinforcement learning framework and the closed-loop control with a potential assistance system.}
    \label{fig:overview_alt}
\end{figure*}

Endovascular interventions are considered the standard of care for a wide range of often critical conditions, including myocardial infarction and larger vessel occlusive ischaemic strokes. 
However, they are not without challenges.
Physicians need to operate alongside the patient and the fluoroscopy equipment, which exposes them to radiation and poses a health risk. Although modern equipment is designed to minimize this exposure, further reductions are desirable \cite{ho_ionizing_2007, bockler_praktische_2020, crinnion_robotics_2022}.
Side-effects from frequently wearing the heavy gear for radiation protection, normally lead-impregnated fabric, can be orthopedic problems \cite{goldstein_occupational_2004}. 
The complexity of endovascular treatments requires extensive specialized training and equipment. 
This leads to a potential scarcity of proficient physicians, particularly in remote and underserved regions \cite{yan_access_2022}. 

Robot-assisted endovascular interventions offer a promising solution to address these challenges faced by physicians \cite{zhao_remote_2022,mitros_theoretical_2022}. 
These robots, with their remote manipulation capabilities, allow operators to be shielded from radiation, enabling patient positioning that optimizes radiation exposure regardless of ergonomic constraints \cite{patel_comparison_2020}. 
Additionally, they facilitate remote interventions via telesurgery \cite{patel_long_2019}.

The next step in the evolution of endovascular robots involves integrating assistance systems and higher levels of automation to support surgeons during procedures \cite{duan_technical_2023}.
The potential benefits of automation in interventions include enhanced patient safety and quality of care, as well as reduced hospital expenditures \cite{attanasio_autonomy_2021}. 
Autonomous navigation holds significant potential for enhancing patient care by allowing physicians to focus on critical aspects of intervention planning and supervision, rather than the complex task of device manipulation, thereby reducing their workload \cite{pore_autonomous_2023}. 
Moreover, advancements towards autonomous navigation can serve as a foundation for simpler assistance systems, such as training platforms for novice surgeons or partially autonomous systems that suggest optimal movements to the surgeon, either digitally or through haptic feedback, akin to lane-keeping assistance in automobiles.

In recent years, research about autonomous navigation in endovascular interventions has increased in concentration.
Early attempts with conventional control had only limited success \cite{ji_guidewire_2011, jayender_autonomous_2008}. 
However, recent publications have shifted their focus toward learning-based approaches, recognizing that endovascular interventions are particularly well suited for the application of machine learning techniques \cite{li_machine_2022}. 

However, direct comparison of the effectiveness of these different control algorithms is hampered by significant differences in their evaluation environments.  
Simulation-based approaches, for instance, have utilized either a 2D physics \cite{ritter_quality-dependent_2022} or a 3D finite element simulation of realistic vessel systems \cite{meng_evaluation_2022, schegg_automated_2022, karstensen_recurrent_2023, jianu_cathsim_2024} and synthetic vessel systems \cite{kienzlen_concept_2022, scarponi_zero-shot_2024} for evaluation.  

In physical real-world applications, some studies \cite{zhao_cnn-based_2019, kweon_deep_2021, li_discrete_2023, li_casog_2024} use phantoms with technically motivated geometries constructed from transparent hard plastic where feedback is received through an overhead-mounted camera. Others employ flexible phantoms replicating the vessel geometries of real patients, with feedback derived from camera images \cite{wang_study_2022, yang_guidewire_2022, song_learning-based_2022} or an electromagnetic tracking system \cite{chi_collaborative_2020}. 
In another study \cite{karstensen_learning-based_2022}, evaluation was conducted in ex-vivo animal tissue, utilizing fluoroscopy as feedback.

Each study is conducted in its own unique evaluation environment, featuring vessel models with specific geometries and material characteristics. 
Additionally, the absence of open-source availability for both training scripts and evaluation setups poses a significant obstacle to reproducibility \cite{robertshaw_artificial_2023}.

Standardized benchmark interventions are imperative to enhance comparability in autonomous navigation in robotic endovascular interventions. 
If open-source available, these benchmarks can foster research similarly to open-source datasets for vision based assistance systems \cite{hong_cholecseg8k_2020} or the Farama gymnasium environments for reinforcement learning algorithms \cite{towers_gymnasium_2023}.
Commercial simulators such as the VIST\textsuperscript{\textregistered} system (Mentice AB, Sweden) or CATHIS\textsuperscript{\textregistered} system (CATHI GmbH, Germany) could serve as benchmarks, but while these systems offer cutting-edge simulation capabilities, they are proprietary solutions lacking the depth of access and availability required for rigorous research purposes. 

To provide a common simulation-based environment for evaluation and enhance comparability across studies we propose three distinct endovascular benchmark interventions.
These digital benchmarks are easily adoptable by researchers and have the potential to drive the development of new control paradigms.
The benchmarks are based on \textit{stEVE} (\textit{\textbf{s}imula\textbf{t}ed \textbf{E}ndo\textbf{V}ascular \textbf{E}nvironment}), a modular simulation framework tailored for endovascular interventions with the aim of catalysing the further development of assistive systems, focusing on, but not limited to, reinforcement learning.

Furthermore, we performed experiments exhibiting the real-world applicability of the simulated benchmarks and provide baseline results for autonomous device control based on our methodology from \cite{karstensen_recurrent_2023}. 

\begin{table*}[t]
\centering
\caption{Comparison of simulation frameworks in autonomous endovascular navigation research.}
\label{tab:sim_models}
\begin{tabular}{lccccc}
\hline
Research        & Engine & Devices  & Vessel system         & Open Source & Flexibility \\
\hline
Molinero et al. \cite{molinero_haptic_2019} & Unity             & catheter   & single (aorta)             &  -         & -          \\
Meng et al. \cite{meng_evaluation_2022}    & Unity             & guidewire   & single (aorta)             & -          & -          \\
Schegg et al. \cite{schegg_automated_2022}  & SOFA              & guidewire  & various (coronary)             &  -         & -          \\
Kienzlen et al. \cite{kienzlen_concept_2022} & SOFA              & guidewire & single (synthetic)              &  -         & -          \\
Scarponi et al. \cite{scarponi_zero-shot_2024} & SOFA              & guidewire  & various (synthetic)             &  -         & -          \\
Jianu et al. \cite{jianu_cathsim_2024}    & MuJoCo            & catheter   & single (aorta)             & \checkmark         & low        \\
\hline
\textbf{stEVE} (ours)            & SOFA              & guidewire \& catheters & various (aorta \& cerebral)  & \checkmark         & high      \\
\hline
\end{tabular}

\end{table*}

\noindent The key contributions are summarized as follows:
\begin{enumerate}[label=(\roman*)]
    \item Autonomous control algorithm and training algorithm demonstrating successful simulation-to-reality transitions, robust patient-individual generalization, and initial experiments in autonomous dual-device navigation.%
    \footnote[1]{See \href{https://github.com/lkarstensen/stEVE_training}{https://github.com/lkarstensen/stEVE\_training}.\label{fn_eve_training}}
    \item Introduction of benchmark interventions for autonomous endovascular navigation to improve comparability and lower the barrier to entry for future studies.%
    \footnote{See \href{https://github.com/lkarstensen/stEVE_bench}{https://github.com/lkarstensen/stEVE\_bench}.\label{fn_eve_bench}}
    \item Proposition of the modular, open-source framework \textit{stEVE} to advance research in assistance systems for robotic endovascular interventions.%
    \footnote{See \href{https://github.com/lkarstensen/stEVE}{https://github.com/lkarstensen/stEVE}.\label{fn_eve}}
\end{enumerate}

We have chosen a modular approach to ensure flexible use. Our training scripts\footref{fn_eve_training} facilitate the reproduction of our results and provide a low entry barrier for other researchers to set up their own training. 
By separating the benchmark environments\footref{fn_eve_bench} from the training, researchers can implement their own training routines using the reinforcement learning framework of their choice.

Additionally, the \textit{stEVE} simulation framework can be utilized for other research in endovascular interventions beyond the scope of autonomous navigation.

\section{stEVE - simulated EndoVascular Environment}

The software framework stEVE offers a modular toolkit which enables users to flexibly compose simulated interventions according to their clinical needs and technical requirements. 
Furthermore, the intervention data can be seamlessly converted to feedback suitable for reinforcement learning.
An overview of the components of stEVE, their existing implementations and the embedding with an assistance system is shown in Fig.~\ref{fig:overview_alt}.

An overview of simulation frameworks for research in autonomous endovascular navigation is given in Table \ref{tab:sim_models}. In comparison, stEVE offers a wider range of devices and vessel systems, is open source available, and exhibits a higher degree of modularity and extendability. Especially the last point is an advantage in the adoption of this framework for future research. 

In the subsequent sections, we explore the medical background, derive the components of a simulated intervention, and explore the reinforcement learning interface. 

\subsection{Medical Background}

During endovascular procedures, devices are typically introduced into the vascular system at specific sites, including the femoral artery (located in the groin), the brachial artery (at the elbow), or the radial artery (at the wrist). 
Subsequently, these devices must be skillfully navigated through the vascular network by probing the correct vessel at bifurcations to the target region for further treatment, e.g., mechanical thrombectomy in cerebral vessels in the context of stroke. 
This navigation process is achieved by external manipulation of the distal end of the device to facilitate entry of the proximal end into the target artery. \cite{brilakis_chapter_2021}

In many cases, simultaneous manipulation of two concentric devices (e.g., guidewire and catheter) becomes necessary, as they mutually influence the combined stiffness and shape of the tip of the devices utilized for navigation. 
While sometimes several devices are inserted in parallel at the same insertion point, e.g., for fenestrated endovascular aortic aneurysm repair, or at different insertion points, e.g., for treatment of aortic bifurcation stenoses, most interventions utilize a concentric arrangement of devices at one insertion point. \cite{brilakis_chapter_2021-1}

Throughout this intricate navigation procedure, continuous feedback is provided through fluoroscopy imaging, a low-dose x-ray modality that provides real-time visualization of the device positions inside the vasculature.
When the devices reach the target point, the navigation is successful and the treatment of the lesion can commence. 
In this work, we consider the navigation aspect of the intervention. \cite{schneider_endovascular_2019}

\subsection{Components of the Intervention within stEVE}
stEVE comprises simulated interventions consisting of the vessel tree including an insertion point, the target site in the vasculature, the utilized devices, fluoroscopic imaging and a simulation engine. 

\subsubsection{Vessel tree with insertion point} 
Considering the unique vessel geometry of each patient, it is essential that the vessel tree is designed for easy adaptation (e.g., patient-specific anatomy, pathologies) or replacement as required.
Moreover, physicians determine the optimal access site based on the planned intervention, the patient anatomy, and the respective target area. Consequently, it is essential that the insertion point can be flexibly defined inside the simulation framework.
The vessel tree defines the geometric characteristics of the vascular system. It provides information about the vessel geometry in the form of an ordered centerline point cloud and a surface mesh representing the vessel walls.
Additionally, an insertion point is specified, consisting of both position and direction. 
Currently, we provide different methods for integrating vessel trees. 
Users can either load a custom vessel mesh from any anatomy, utilize vessel models offered by the vascular model repository \cite{wilson_vascular_2013}, or generate parameterized artificial aortic arches of six different aortic arch types \cite{karstensen_recurrent_2023}.

\subsubsection{Target site} 
For each procedure, there should be a possibility to specify the target point of the navigation and modify it, if required, during the intervention's progression.
Depending on the specific intervention and the patient's vessel geometry, targets can either be manually selected by the user or automatically chosen in the target vessel. 
The target is considered reached when the tip of any device approaches a predefined threshold distance of the target location.

\subsubsection{Devices} 
The devices employed in interventions exhibit diverse mechanical characteristics and shapes. 
The choice of device depends on the vessel geometry, the target within that geometry, and/or the physician's preference.
Although the majority of devices are readily available off-the-shelf, there are instances where the device is customized by the physician, such as coronary guidewires.
Hence, the capability to define arbitrarily shaped devices within the simulation framework is necessary.
stEVE offers the flexibility to create devices of various shapes by connecting straight segments and segments curved in one (i.e., bend) or two (i.e., helix) dimensions. 
Additionally, the framework includes parameterized standard devices such as \textit{straight} or \textit{j-shaped} devices, as well as the \textit{Simmons} catheter \cite{baba_yahya_simmons_2023}.

\subsubsection{Fluoroscopy} 
An adaptable fluoroscopy feedback mechanism is essential, supporting configurations such as single-plane or biplane imaging systems \cite{cowen_cardiovascular_2018}.
stEVE provides the capability to instantiate simulated fluoroscopy systems as needed. 
Depending on the context, multiple systems with different viewing angles can be applied. 
Currently, two types of fluoroscopy feedback are implemented: 
a simulated tracking algorithm with device coordinates as output and a basic fluoroscopy simulator, which takes the devices positions and creates a fluoroscopy-like image by inserting the devices in a noisy gray-scale image.

\subsubsection{Simulation engine}
To simulate an intervention, i.e., device movement within the specified vessel tree, stEVE relies on the simulation engine SOFA \cite{faure_sofa_2012} with the \mbox{BeamAdapter} plugin \cite{wei_near_2012}. 
As seen in Table \ref{tab:sim_models}, SOFA is utilized most often in current research, hinting at its suitability for this task.
This simulation engine utilizes the finite-element-method (FEM) where the vessel wall is modelled as a mesh and devices as a chain of 1-dimensional elements based on Kirchhoff's rod theory. 
Translation and rotation is applied to the device at the insertion point, propagated along the device and a collision model keeps the device inside the vessel structure.

\subsection{Reinforcement Learning Interface}
The prevailing standard interface for reinforcement learning is set by the Farama Gymnasium \cite{towers_gymnasium_2023}, which describes a structured interface that includes data types for

\begin{list}{}{}
    \item \itab{\textit{observations},} \tab{e.g., device and target position,}
    \item \itab{\textit{rewards},} \tab{e.g., $0.001 \cdot \Delta\text{(distance to target)}$,}
    \item \itab{\textit{terminal states},} \tab{e.g., target reached,}
    \item \itab{\textit{truncations},} \tab{e.g., timeout, and}
    \item \itab{\textit{infos},} \tab{e.g., current translation speed.}
\end{list}

stEVE offers a suite of tools designed to facilitate the conversion of intervention data into the Gymnasium interface. 
The interface definition can draw on the full intervention data, i.e., vessel geometry, three-dimensional target position, device shape and characteristics, three-dimensional position data of each device, and fluoroscopy images.

It is imperative for the user to bear in mind the real-world accessibility of data, as access to comprehensive intervention data in the simulation significantly differs from the constraints encountered in a real-world intervention. 
For example, the simulation provides 3D positional information for each device, while from a single fluoroscopy image only 2D positions are available in the real world. 
Furthermore, in cases where control design is exclusively performed in a simulated environment and subsequently evaluated in the real world, the reward function can leverage the data-rich environment of the simulation. 
However, observations must be carefully tailored and adjusted to account for the real-world limitations encountered during the evaluation phase. 

\section{Benchmark Environments}

\begin{figure*}[t]
    \centering
    \subfloat[]{\includegraphics[width=2.77cm]{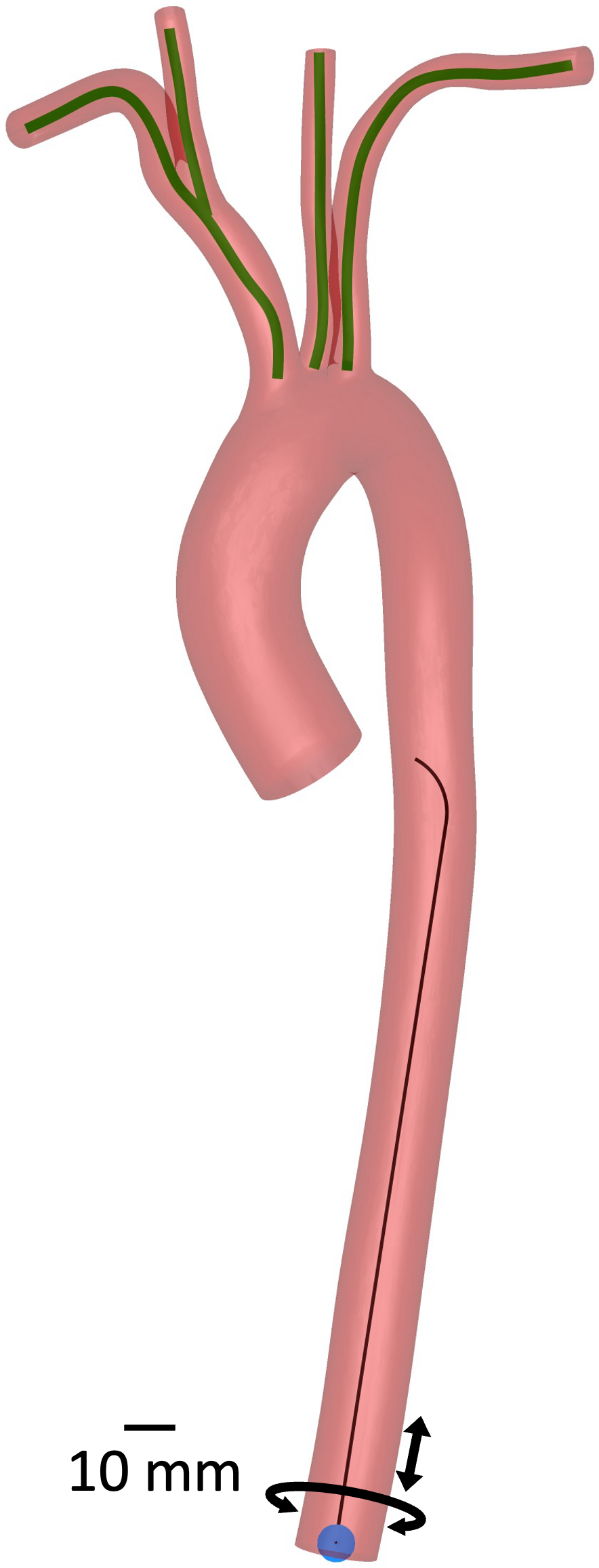}
        \label{fig:benchmark_1}}
    \hspace{0.6cm}
    \rulesep
    \hspace{0.6cm}
    \subfloat[]{\includegraphics[width=6.4cm]{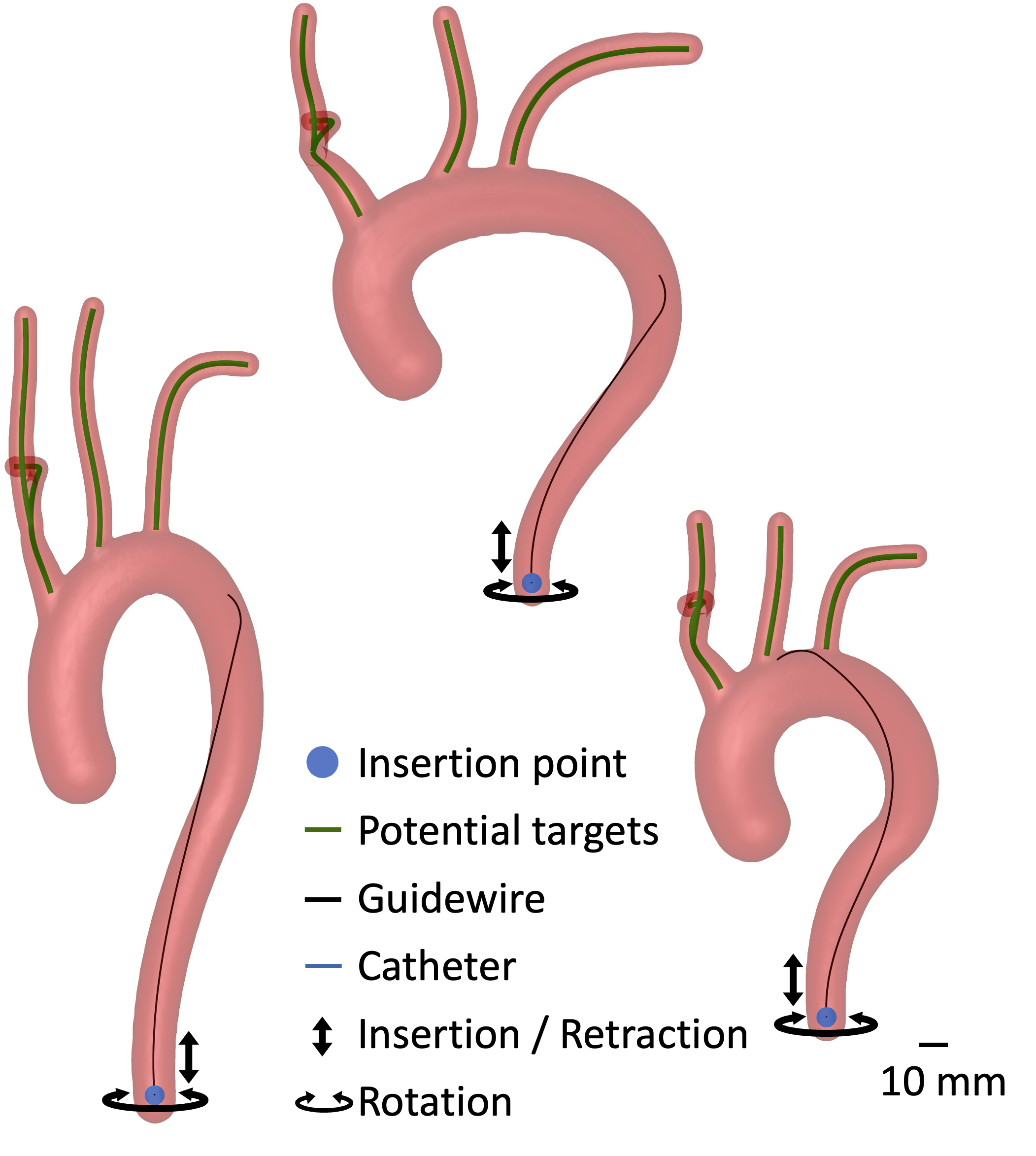}
        \label{fig:benchmark_2}}
    \hspace{0.6cm}
    \rulesep
    \hspace{0.6cm}
    \subfloat[]{\includegraphics[width=2.5cm]{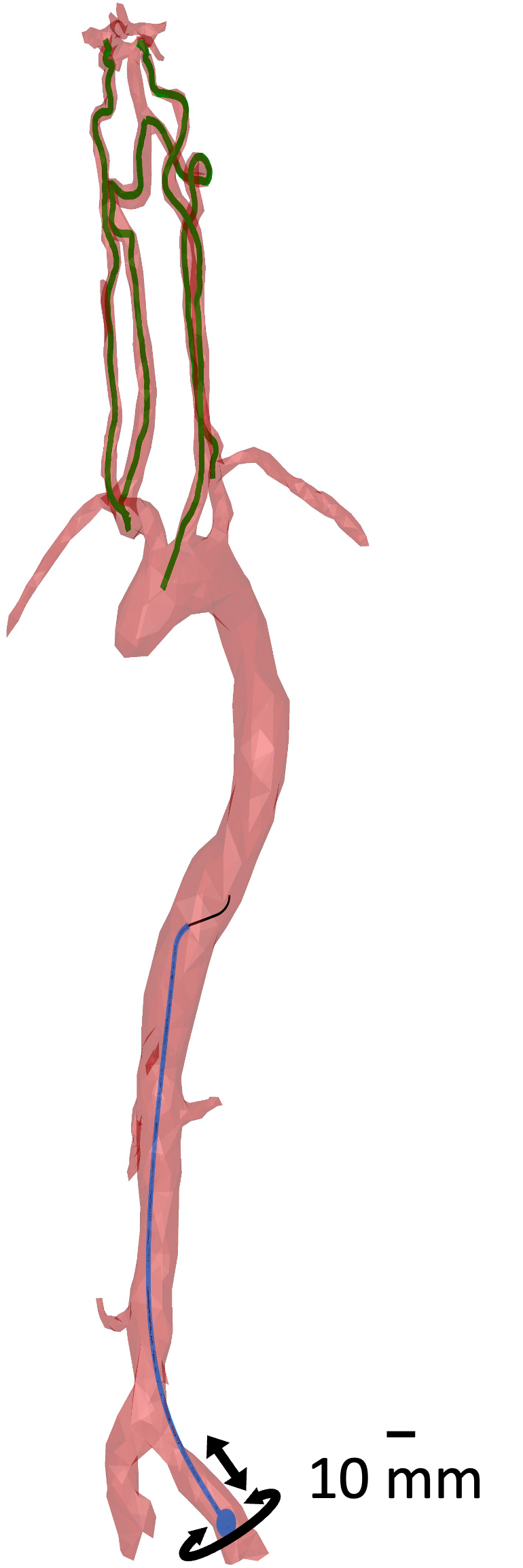}
        \label{fig:benchmark_3}}
    \caption{Proposed benchmark interventions: (a) \textit{BasicWireNav}: single real aortic arch, single device and supra-aortic targets. (b) \textit{ArchVariety}: multiple artificial aortic arches, single device and supra-aortic targets. (c) \textit{DualDeviceNav}: single real vascular system, dual device and targets in the common carotid and vertebral arteries.}
    \label{fig:benchmarks}
\end{figure*}

Based on the medical background, we propose three distinct benchmark environments, utilizing the stEVE framework. Each benchmark focuses on distinct facets of device navigation during endovascular interventions: fundamental guidewire maneuvering, adaptability to patient-specific variations, and concurrent handling of two concentric devices (i.e., guidewire and catheter).

\subsection{BasicWireNav} This benchmark involves navigating a guidewire through the aortic arch to the supra-aortic arteries in a single patient. It assesses fundamental navigation capabilities, i.e., navigating a guidewire to probe the desired artery in a fixed vessel geometry. The vascular model is derived from a healthy 23-year-old female from the vascular model repository (\#0011\_H\_AO\_H) \cite{wilson_vascular_2013}. The navigation target is randomly selected for each episode along the centerline of the brachiocephalic trunk, carotid, or subclavian arteries, resulting in 562 different potential targets. Figure \ref{fig:benchmark_1} depicts the navigation task.

\subsection{ArchVariety} The second benchmark entails guiding a guidewire through the aortic arch to the supra-aortic arteries in the vasculature of patients with varying anatomies. This benchmark evaluates generalization capabilities and adaptability to patient-specific variations. The vascular models consist of randomly generated type I aortic arches. The navigation target is randomly selected for each episode along the centerline of the brachiocephalic trunk, carotid, or subclavian arteries. Figure \ref{fig:benchmark_2} provides a depiction of the navigation task for three different aortic arches.

\subsection{DualDeviceNav} The third benchmark encompasses guiding a concentric combination of guidewire and catheter through the aortic arch to the supra-aortic arteries and along the cerebral artery to a cerebral target in a single patient. This benchmark evaluates the proficiency in navigating two interacting devices. The patient's vascular model is manually segmented from a computed tomography angiographic scan of the aorta and a magnetic resonance angiography scan of the cerebral vessels. Subsequently, these segments were manually assembled into a cohesive model. Navigation targets are randomly placed along the centerline of the common carotid or vertebral arteries, resulting in 898 potential targets. Figure \ref{fig:benchmark_3} shows an illustration of the vessel system and navigation task.

In all three benchmarks the devices are j-shaped with a tip radius of 12.1~mm and a tip angle of $0.4\pi$ radians and velocities are limited to 35 $mm/s$ translational and 3.14 $rad/s$ rotational speed.
Tracking data is utilized as feedback, rather than fluoroscopy images. 
A majority of studies utilize tracking data as input for their control algorithm either with an electromagnetic tracking system or a tracking algorithm analyzing the images \cite{robertshaw_artificial_2023}.
Simulating the tracking algorithm instead of images ensures that the performance of the tracking system does not impact the performance of the control algorithm. 
For this purpose the simulated tracking algorithm extracts the device position from the simulation and transforms them to a coordinate system in an imaginary right anterior oblique projection image plane. 
The imaging frequency is set at 7.5~Hz, a typical value for optimized radiation reduction \cite{abdelaal_effectiveness_2014}.

\section{Autonomous Control}
We performed experiments of autonomous device navigation on the three benchmarks and evaluated in simulation. 
BasicWireNav and ArchVariety controllers are additionally evaluated on physical test benches. 
A separate controller was trained for each benchmark with the same training routine and general architecture.
These experiments show applicability of the benchmarks, transferability from simulation to the real world, and serve as baselines for future comparison of navigation strategies for each benchmark. 
In this section, we detail our controller architecture and approach to training, as well as the evaluation modalities.

\subsection{Controller Architecture and Training Scheme}

The controller applies the soft-actor-critic (SAC) method \cite{haarnoja_soft_2018} and is based on the procedure outlined in \cite{karstensen_recurrent_2023}. 

\begin{figure}[t]
\centering
\includegraphics[width=\linewidth]{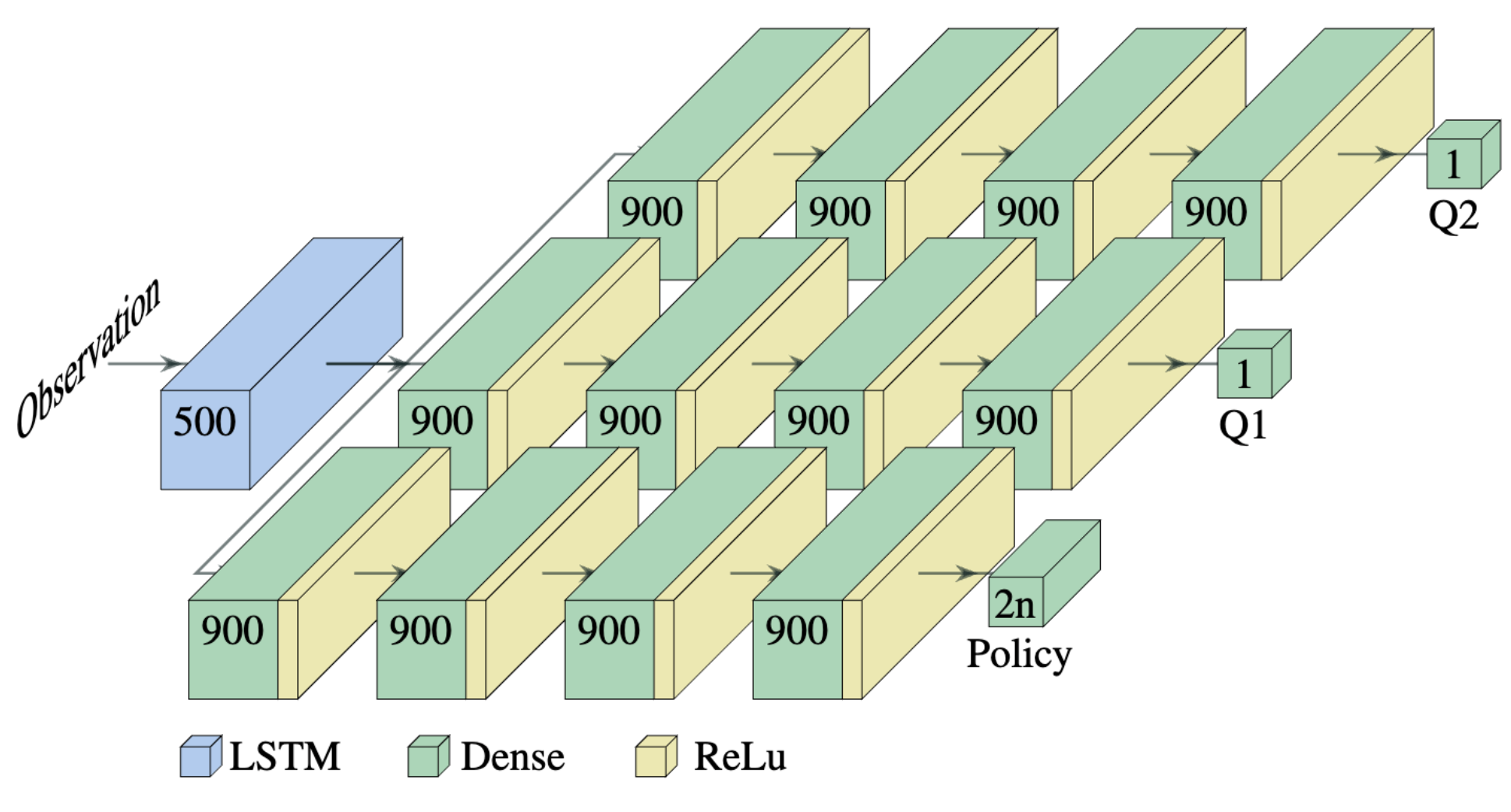}
\caption{The structure of the neural networks of the soft-actor-critic agent with input embedder, Q1- and Q2-network with one output node and policy-network with $2n$ output nodes, where n is the amount of devices. Amount of layers and nodes is exemplary for the \textit{BasicWireNav} controller.}
\label{fig:nn_structure}
\end{figure}

The structure of the neural networks is illustrated in \mbox{Fig. \ref{fig:nn_structure}}. 
The observation is processed by an observation embedder of long short-term memory (LSTM) nodes without activation function. 
The embedded observation is subsequently fed to the policy-, Q1- and Q2-network. 
Each of them consists of strictly feed-forward densely connected hidden layers with a rectified linear unit (ReLu) as activation function. 
For the Q1- and Q2-network the output layer has 1 neuron representing the Q-value, which is solely utilized for training of the neural network.
The output layer of the policy-network has $2 \times n_{devices}$ neurons representing the mean value and standard deviation of the translation and rotation speed for each device. 
The illustrated number of layers and nodes for the observation embedder and hidden layers are exemplary for the \textit{BasicWireNav} controller.

The observation provided to the controller is comprised of four components: the current and previous device tip position, target position, and the last executed action. 
The tip position is defined by three points equally spaced 2mm apart along the device starting from the device tip. 
Each point of the device tip, as well as the target position is given as two-dimensional coordinates in the image plane of the imaging-system. 
The action is provided as translational and rotational speed at the insertion point for each device. 
Each observation is normalized to the range $[-1, 1]$ where -1 corresponds to the minimal and 1 to the maximal value of the observation element in the navigation task.

The reward (R) per step is defined as follows:
\begin{equation*}
R = -0.005 + 0.001\cdot\Delta \textit{pathlength} + \begin{cases}
    1.0& \text{if target reached.}\\
    0  & \text{otherwise.}
\end{cases}
\end{equation*}
where \textit{pathlength} represents the spatial distance between the device tip and the target along the vessel centerlines in millimeters. 

An episode corresponds to one navigation task from the insertion point to the target and is considered successful if the target is reached within a 5\,mm threshold. 
The episode is deemed unsuccessful if a time limit is reached, the device tip reaches the end of the incorrect vessel branch, or a simulation error occurs. 
The control frequency matches the imaging frequency. 

Training is solely performed in simulation and is carried out for $2 \times 10^7$ exploration steps, i.e., cycles of the control loop, as preliminary experiments have shown that this is more than sufficient for the controller to reach its maximal success rate. 
To monitor training progress, the controller conducts 100 evaluation episodes every $2.5 \times 10^5$ exploration steps.

\begin{figure*}[t]
    \centering
    \subfloat[]{\includegraphics[width=8.5cm]{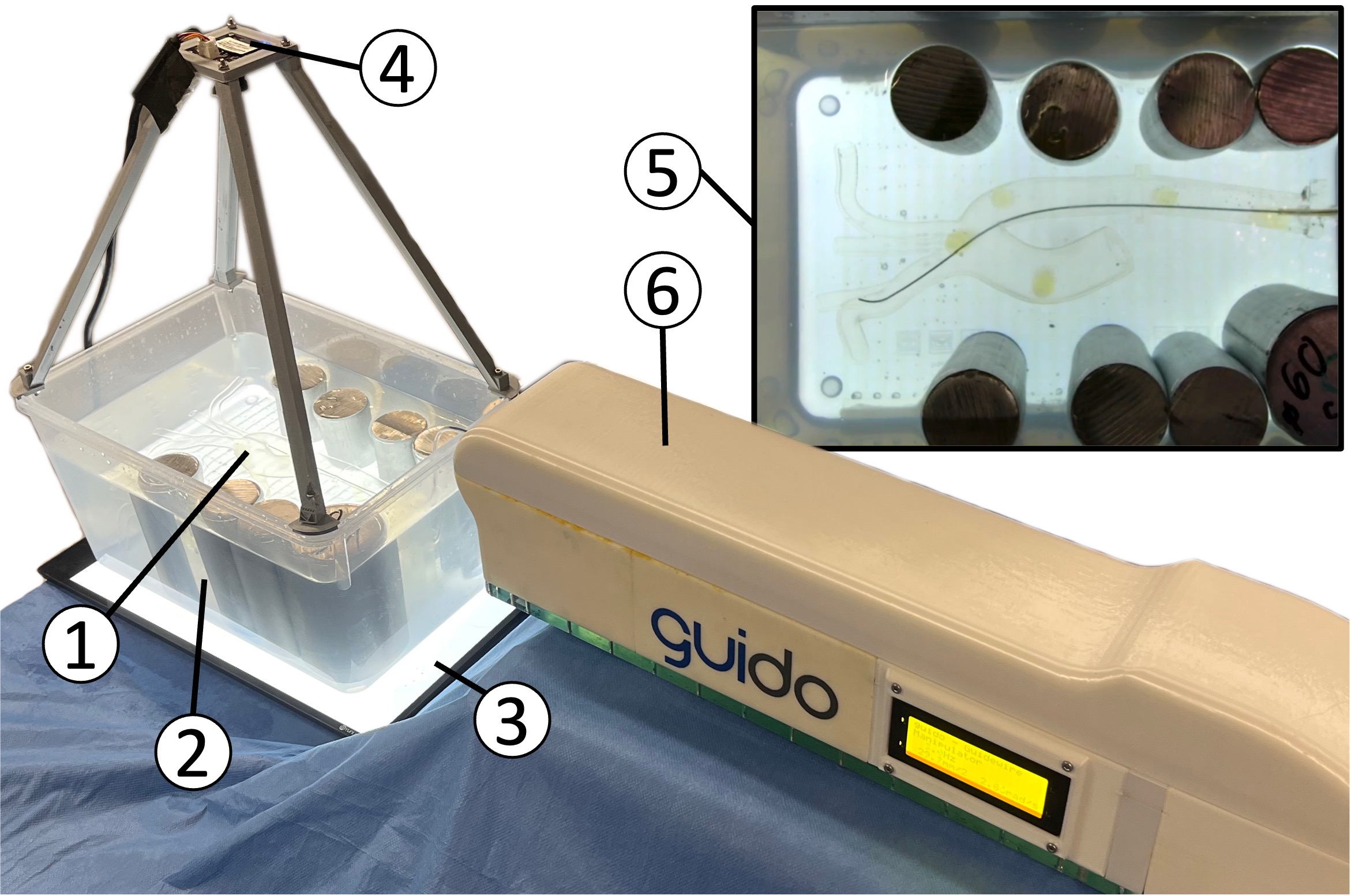}
        \label{fig:testbench_camera}}
    \hfill
    \subfloat[]{\includegraphics[width=8.5cm]{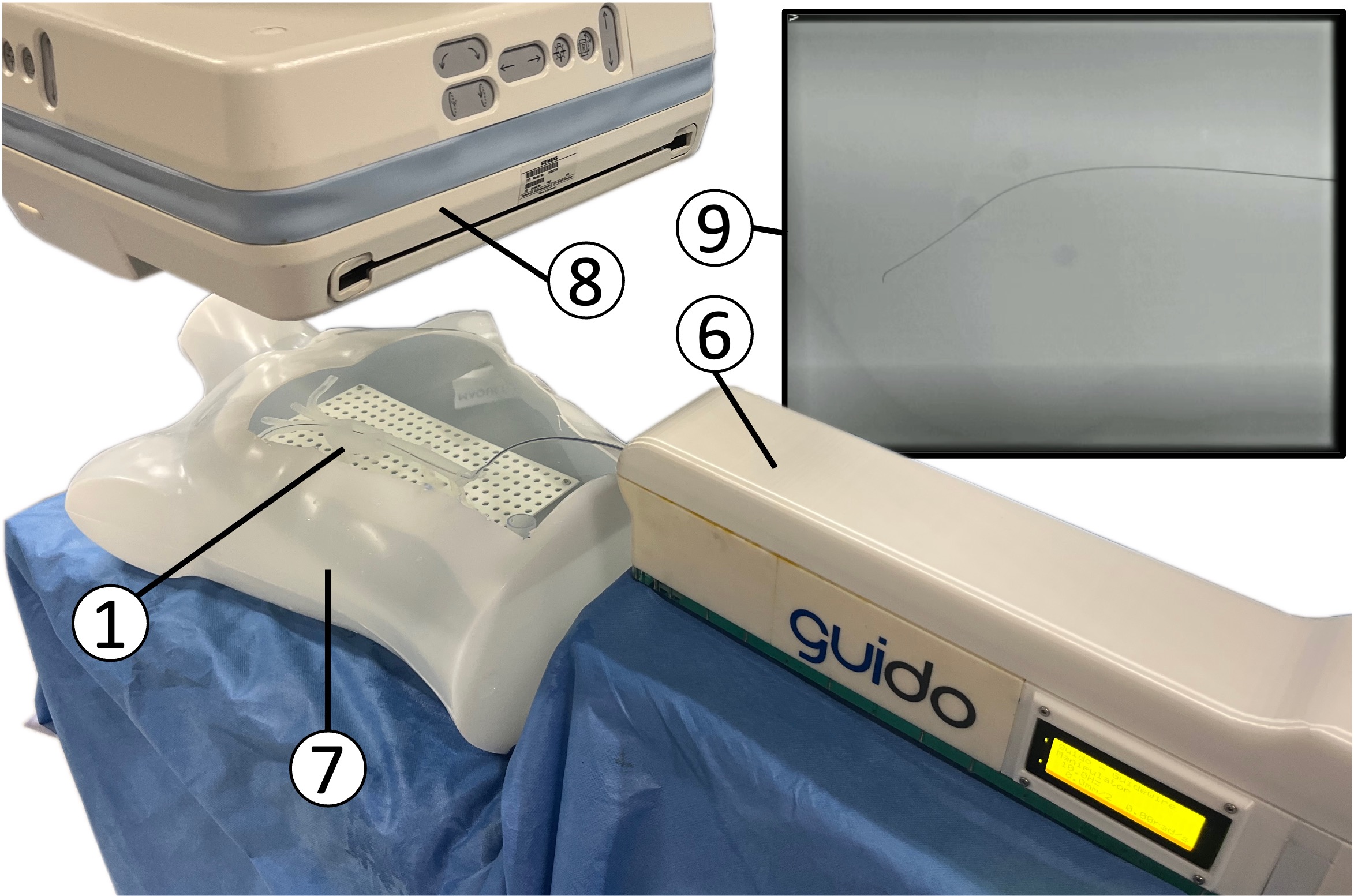}
        \label{fig:testbench_x_ray}}
    \caption{(a) Test bench with camera feedback in phantom \textcircled{\raisebox{-0.9pt}{1}} in a specimen container filled with glycerin \textcircled{\raisebox{-0.9pt}{2}} on a light source \textcircled{\raisebox{-0.9pt}{3}}, a camera mounted above \textcircled{\raisebox{-0.9pt}{4}} which generates real images \textcircled{\raisebox{-0.9pt}{5}}. Images are processed by a tracking algorithm and the controller calculates new actions based on device tracking (not depicted). Actions are performed by a guidewire manipulator \textcircled{\raisebox{-0.9pt}{6}}. \linebreak (b) Test bench with fluoroscopy system \textcircled{\raisebox{-0.9pt}{8}} generating real images \textcircled{\raisebox{-0.9pt}{9}} and water-filled mannequin as specimen container \textcircled{\raisebox{-0.9pt}{7}}.}
    \label{fig:testbenches}
\end{figure*}

\subsection{Evaluation}

We employ three distinct evaluation modalities: (i) in simulation, (ii) a test bench with camera-based feedback (TC), (iii) and a test bench with fluoroscopic feedback (TF).

TC, as depicted in Figure \ref{fig:testbench_camera}, incorporates a 3D printed transparent phantom (Clear, Formlabs Inc., Somerville, USA) submerged in glycerin.
Glycerin is utilized to improve visibility of the device inside the phantom due to the similarity of its refractive index to that of the print resin. 
An overhead camera (SUSB1080P01-LC1100, Ailipu Technology Co., Shenzhen, China) and a lower-mounted light source provide feedback. 
A tracking algorithm is utilized to extract device positions from the camera image \cite{eyberg_ros2-based_2022}. 
A custom guidewire manipulator executes movements (i.e., translation and rotation) to the guidewire (Radifocus Guide Wire M Stiff Type, Angled, 0.035", Terumo Corporation, Tokyo, Japan) instructed by the controller. 
The manipulator has a stepper motor to which the guide wire is directly attached for infinite rotation. 
Guidewire and stepper motor are mounted on a linear rail driven by a second stepper motor, allowing 300 mm of translational movement.
 
TF, as shown in Figure \ref{fig:testbench_x_ray}, mirrors TC, with the phantom, tracking algorithm, manipulator and guidewire being identical. However, in this case, the phantom is placed within a water-filled mannequin, which mimics uniform surrounding tissue and a medical fluoroscopic system (Artis Zeego, Siemens Healthcare GmbH, Erlangen, Germany) provides image-feedback during the intervention.

TC facilitates extensive experimentation on the test bench owing to the accessibility and user-friendly nature of its components. 
TF adds an extra layer of realism by incorporating fluoroscopic feedback. 
The variance in imaging modalities introduces discrepancies in tracking accuracy, primarily attributable to distortion effects, particularly noticeable for movements perpendicular to the image plane.

For \textit{BasicWireNav} training episodes are capped at 20 seconds. This duration was determined to be sufficient for navigation tasks within the velocity constraints, based on manual pre-trials. 
Evaluation episodes extend to 120 seconds, as the method of short training and long evaluation episodes facilitates quicker learning during training and allows ample time for rectifying navigational errors during evaluation. 
Training is performed with an observation embedder of 500 neurons, four layers of 900 densely connected neurons for Q1-, Q2-, and policy-network and a learning rate of $2.199 \times 10^{-4}$.
Evaluations are carried out in simulation and on the TC for 100 targets, with the initial 10 out of 100 targets being additionally assessed on the TF. 
Evaluation on TC examines simulation to reality transfer and the limited scope of evaluation on TF, focusing on just 10 targets, is designed to examine the transition from camera-based to fluoroscopy-based feedback.

For \textit{ArchVariety}, episodes follow the same time limits as \textit{BasicWireNav}. 
Training is performed with an observation embedder of 900 neurons, three layers of 400 densely connected neurons for Q1-, Q2-, and policy-network and a learning rate of $3.218 \times 10^{-4}$.
For evaluation, the aortic arch with seed 661023725 is sampled and 100 targets are navigated both in simulation and on TC. This evaluation setup is designed to assess the simulation-to-reality transfer for a different scenario.

For \textit{DualDeviceNav}, training episodes are limited to 66 seconds, while evaluation episodes last 133 seconds. 
Training is performed with an observation embedder of 500 neurons, four layers of 900 densely connected neurons for Q1-, Q2-, and policy-network and a learning rate of $2.199 \times 10^{-4}$.
The evaluation is carried out exclusively in simulation for 100 targets, offering a controlled and stable environment ideal for benchmarking purposes.

\section{Results}

\begin{table*}[t]
\centering
\caption{Results of the Benchmark Experiments for 100 trials in simulation. \textit{BasicWireNav} additionally evaluated for 10 trials on the camera (TC) and fluoroscopic (TF) test benches.}
\label{tab:results}
\footnotesize
\begin{tabular}{| c | c | c | c | c |}
\hline
\multirow{2}*{Benchmark} & \multirow{2}*{Evaluation} & \multirow{2}*{Success rate}  & Path Ratio [\%] & Duration [s] \\ 
& & & \textit{failed} & \textit{successful} \\

\hline
\hline

\multirow{2}*{BasicWireNav} & Simulation & 98/100 \hspace{2.5mm} \textcolor{white}{10/10} & 94.3 \hspace{2.5mm} \textcolor{white}{00.0} & \textcolor{white}{1}9.6 \hspace{2.5mm} \textcolor{white}{x9.8}\\ 
\cline{2-5}

 & TC & 97/100 \hspace{2.5mm} \textcolor{white}{x}9/10 & 81.6 \hspace{2.5mm} 87.4 & 13.6 \hspace{2.5mm} 18.1\\ 
\cline{2-5}

 \textit{100 trials \hspace{2mm} 10 trials} & TF & \textcolor{white}{00/100} \hspace{2.5mm} 10/10 & \textcolor{white}{00.0} \hspace{2.5mm} \textcolor{white}{00}-\textcolor{white}{0}  & \textcolor{white}{00.0} \hspace{2.5mm} 19.1\\ 
\hline
\hline
\multirow{2}*{ArchVariety} & Simulation & 90/100 & 73.5 & 19.3 \\ 
\cline{2-5}

 & TC & 84/100  & 75.9  & 22.0 \\ 
\hline
\hline

 DualDeviceNav & Simulation & 40/100 & 67.8 & 45.8\\ 
\hline

\end{tabular}
\end{table*}

Table \ref{tab:results} presents the outcomes of the evaluation experiments, with the key metrics \textit{Success Rate}, which quantifies the number of achieved targets out of the attempted ones; 
\textit{Path Ratio}, a measure of the remaining distance to the target in unsuccessful episodes, calculated by dividing the remaining distance by the initial distance; 
and \textit{Duration} of the successful episodes, measuring the time from the start of navigation until the target is reached.

In the \textit{BasicWireNav} experiments result metrics are given for 100 targets for the simulation and TC experiments. Additionally, results metrics are provided for the first 10 of the 100 targets for TC and TF to enable analysis of camera to fluoroscopy transfer on the physical test benches. 

During training the controller achieved its maximum success rate after 46600 episodes, which was achieved after 6h39min with 55 worker agents and one trainer agent on a system equipped with an AMD Threadripper 3970X CPU, NVIDIA RTX 3090 GPU and 128GB RAM. 

For the full 100 targets, success rates reach 98 out of 100 in simulation and 97 out of 100 on TC (p-value 0.653). The path ratio exhibits a significant decline from 94.3\% in the simulated environment to 81.6\% on TC (p-value 0.045). The duration of successful episodes increases from 9.6~s in simulation to 13.6~s on TC (p-value 0.011). 

For the reduced set of 10 targets, the autonomous navigation reaches 9 out of 10 targets on TC and 10 out of 10 targets on TF (p-value 0.583), respectively. The path ratio cannot be calculated for TF, as no episodes resulted in failure. On TC the failed attempt navigated 87.4\% of the path. The duration insignificantly increased from 18.1~s on TC to 19.1~s on TF (p-value 0.323). 

\begin{figure}[t]
    \centering
    \subfloat[]{\includegraphics[width=6cm]{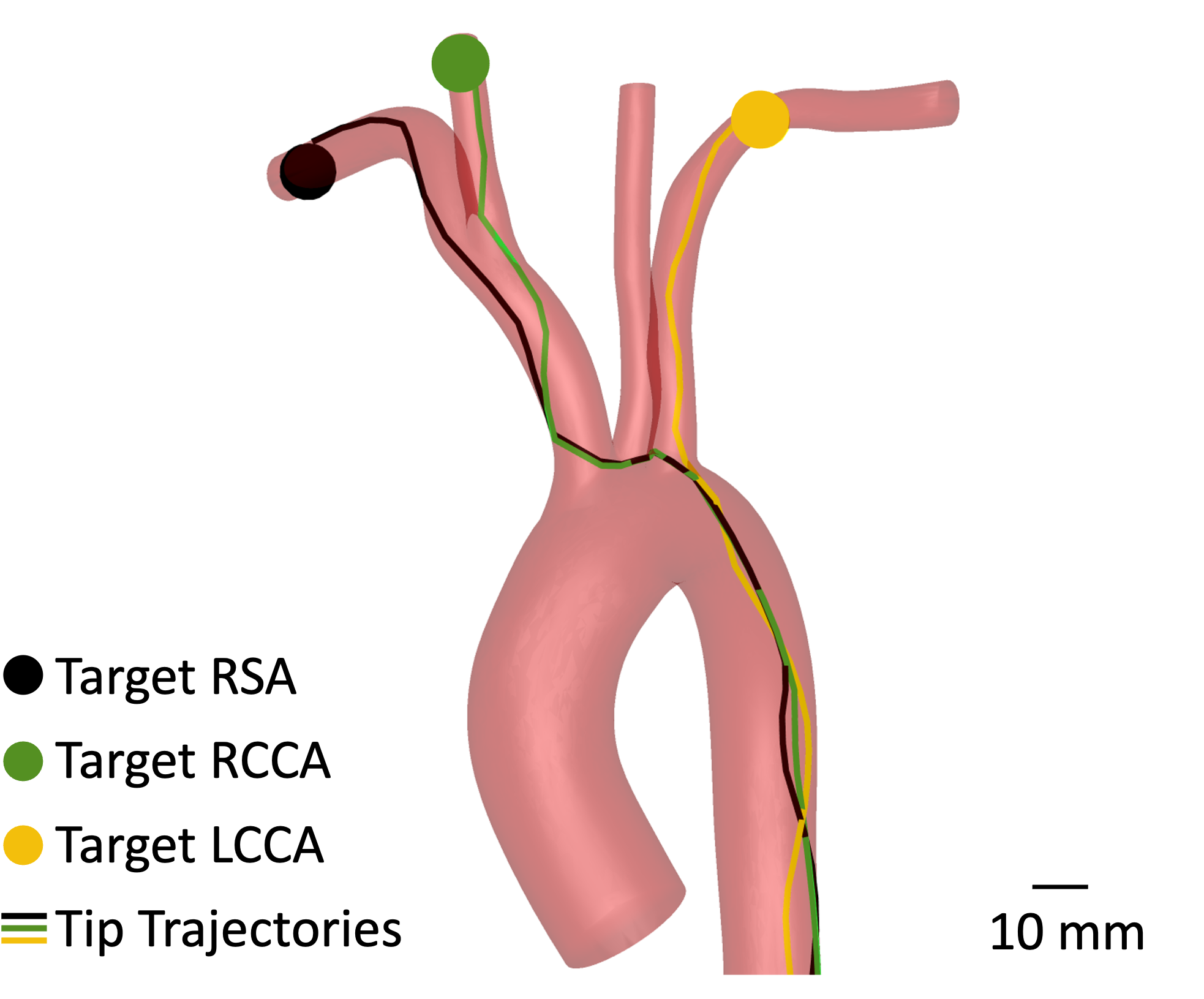}
        \label{fig:trajectory_simulation}}
    \hfill
    \subfloat[]{\includegraphics[width=3.64cm]{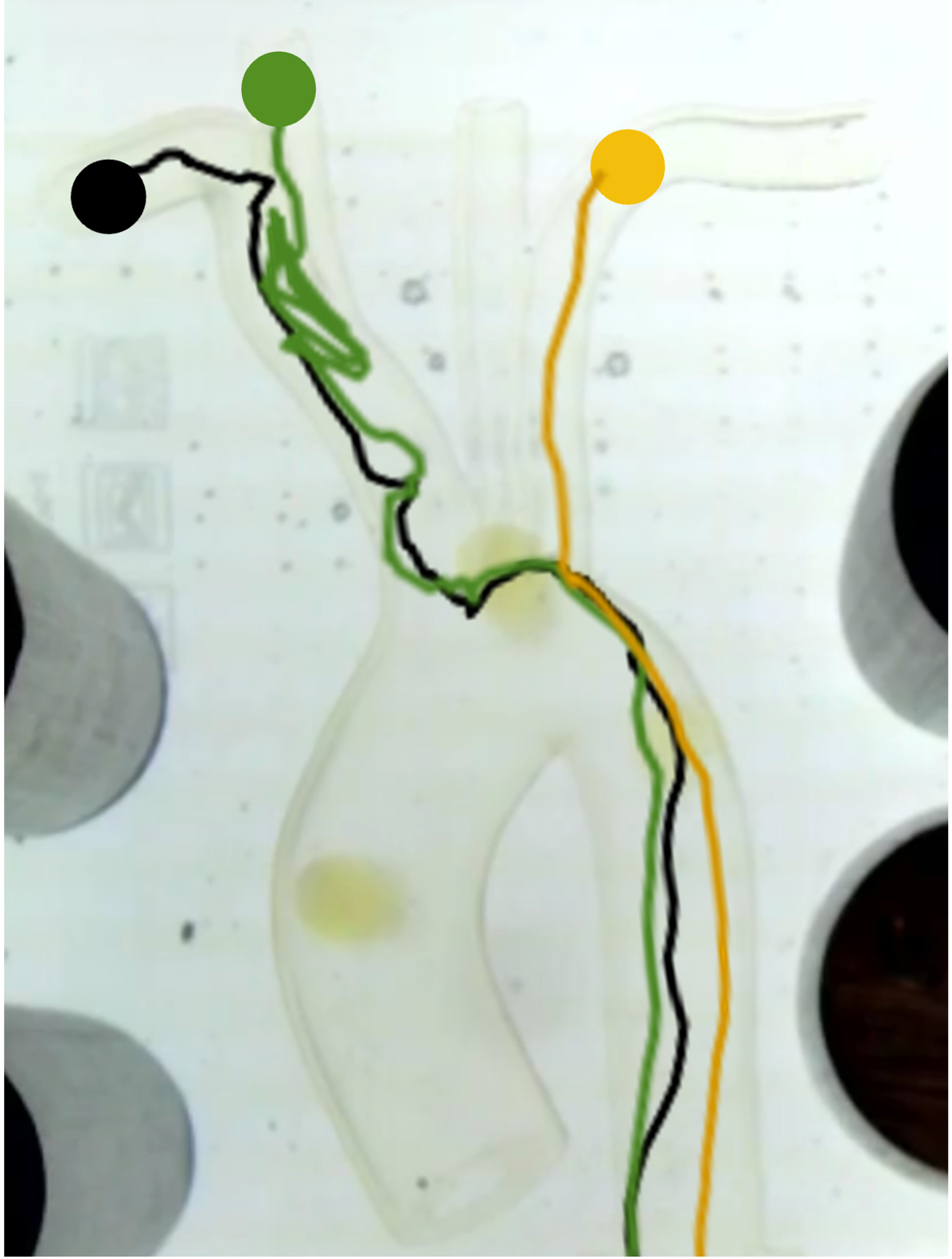}
        \label{fig:trajectory_camera}}
    \hfill
    \subfloat[]{\includegraphics[width=3.64cm]{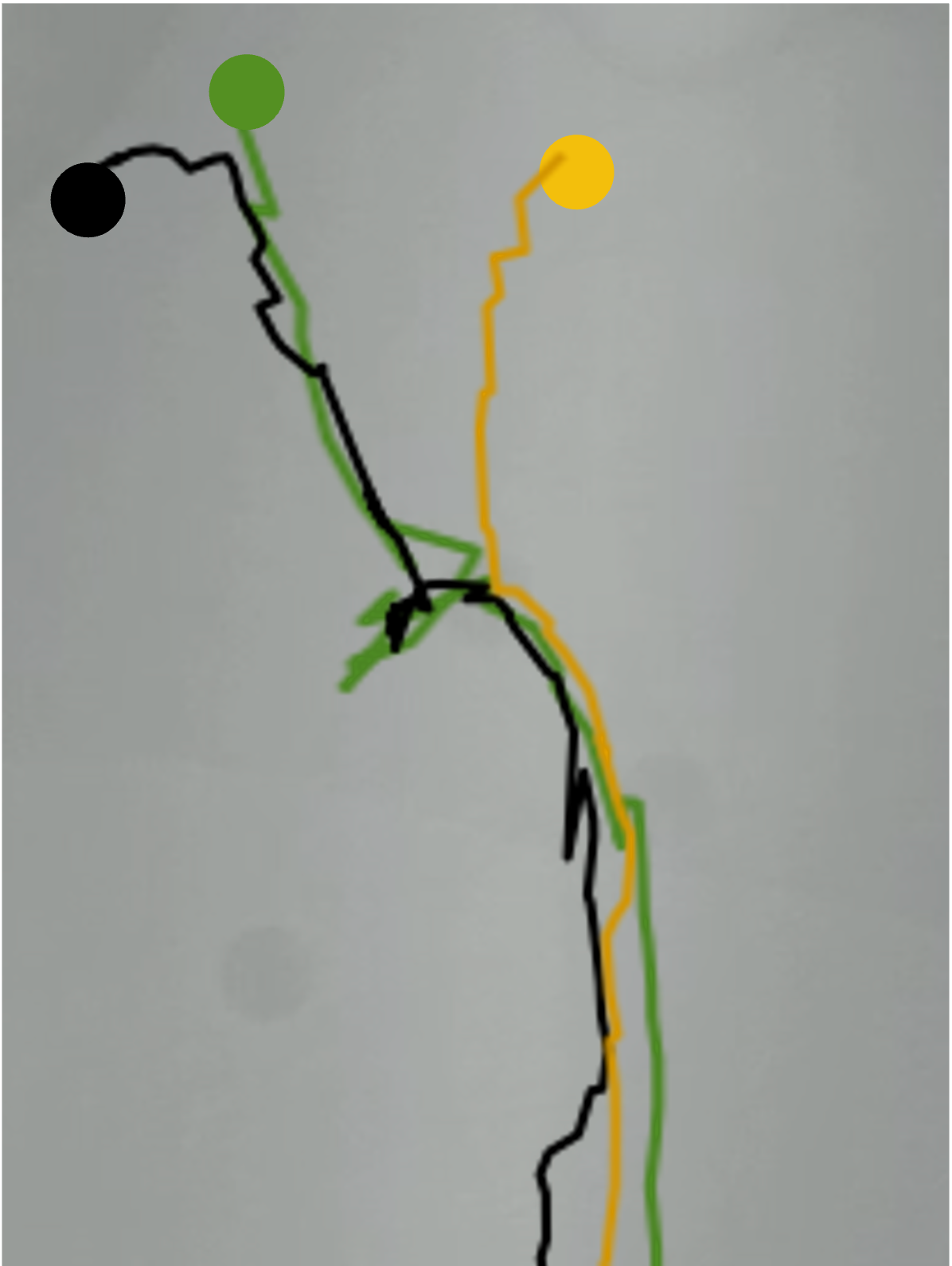}
        \label{fig:trajectory_xray}}
    \caption{Trajectories of \textit{BasicWireNav} navigations to targets in the right subclavian artery (RSA) and right and left common carotid artery (RCCA and LCCA) for (a) simulation, (b) TC and (c) TF.}
    \label{fig:trajectories}
\end{figure}

Trajectories leading to targets in the right subclavian artery (RSA) and right and left common carotid arteries (RCCA and LCCA) are depicted in Fig.~\ref{fig:trajectories} for simulation, TC, and TF. 
In simulation, the three targets are successfully reached without complications within a maximum duration of 11\,s. 
On the test benches, the RSA target is reached on the initial attempt, while for the RCCA and LCCA targets, up to 82\,s of attempting to probe the desired artery were required to access it successfully.

In the \textit{ArchVariety} experiments, the controller reached the maximum success rate after 33700 episodes of training. 
Training lasted 10h40min with 29 worker agents and one trainer agent on a system equipped with a AMD Ryzen 5950X CPU, NVIDIA RTX 3070 Ti GPU and 64GB RAM. 
In the evaluation, a success rate of 90 out of 100 is achieved in simulation, slightly reduced to 84 out of 100 on the camera-based test bench (p-value 0.209). The path ratio remains similar, at 73.5\% and 75.9\%, respectively (p-value 0.713), while the duration increases from 19.3\,s in simulation to 22.0\,s on the test bench (p-value 0.409).

During the \textit{DualDeviceNav} training, the controller reached the maximum success rate after 40800 episodes. 
Training lasted 71h17min with 16 worker agents and one trainer agent on a system equipped with a AMD Ryzen 7950X CPU, NVIDIA RTX 4090 GPU and 64GB RAM. 
The experiments attain a success rate of 40 out of 100, with a path ratio of 67.8\% for the unsuccessful episodes and a duration of 45.8\,s for the successful episodes.

\section{Discussion}

In this paper, we showcased the modular framework \textit{stEVE} to compose simulated endovascular environments and its capability to represent a diverse range of interventions, introduced three benchmark environments as a crucial first step in rendering research and development of autonomous endovascular navigation more comparable and performed experiments in simulation and on real test benches.

The experiments demonstrate transferability from the simulated benchmarks to the real test bench. 
Notably, we only observe an insignificant decline in success rate after simulation-to-reality transfer, but increased episode duration for \textit{BasicWireNav}, presumably attributed to tracking discrepancies between the test bench setup and its simulation. 
An analysis of the trajectories supports this theory as repeated artery probing attempts are only required on the test benches, whereas in the simulation arteries are probed on the first attempt. 
On the one hand this underlines the robustness of the control algorithm and on the other hand shows potential to improve realism of the simulation. 
In simulations, the tracking coordinates are accurate, but a tracking algorithm inherently has noise and inaccuracies when tracking a real device. 
Real images, captured via a point source, include distortions when vessel structures span varying image depths. 
Integrating these effects within the simulation in future work might enhance the outcomes.
Alternatively, exposing the controller to varying vessel geometries during training enhances robustness in the transfer from simulation to reality, as the significant increase in duration cannot be observed in the test bench experiments of \textit{ArchVariety}.

The benchmark environments are a crucial step in making research and development of autonomous endovascular navigation more comparable. 
As demonstrated in the experiments, these benchmarks are suitable for learning-based control schemes and serve as basis for further research aimed at improving control algorithms.
They include important skills for the navigation of endovascular devices.

Benchmark \textit{BasicWireNav} checks the basic manipulation of a guidewire and our controller showed the ability to solve this task in simulation as well as in the real world.
Benchmark \textit{ArchVariety} checks patient individual adaptability of the vessel geometry. 
Our controller demonstrated favorable results with the given training procedure but still leaves room for refinement. 
Future enhancements could involve the inclusion of features specifically tailored for generalization, for instance, geometry information from pre-interventional imaging.
Benchmark \textit{DualDeviceNav} addresses the challenging problem of concurrent manipulation of two concentric devices over a long navigation route. 
The experiment resulted in a moderate but promising success rate of 40 out of 100 and shows the need for further research.
Qualitative analysis of navigation attempts revealed that the device inserted further exhibited expected movements, while the other device displayed chaotic motion. 
This discrepancy can be attributed to the observation’s focus solely on the tip of the more deeply inserted device, neglecting information about the shorter one.
To enhance performance, potential methods include integrating additional information about individual devices into the observation, such as the insertion length, or refining the reward system to penalize unfavorable relative positions.
Additionally, enhancing the underlying reinforcement learning method, such as adopting a multi-agent approach, could prove beneficial.

However, it is important to note that while the presented benchmarks extend the current state of the art, they are limited and do not fully encompass the complexity of real interventions. 
For instance, provided benchmarks do not yet account for the device selection process, intra-interventional device exchanges, or the repositioning of the imaging system. 
Furthermore, benchmarks giving simulated fluoroscopy images as feedback are missing and should be implemented in the future to provide an opportunity to compare control algorithms in an end-to-end manner. 

The framework \textit{stEVE} offers a modular approach to compose endovascular interventions and embed them in reinforcement learning environments.
Its ease of adaptation for individualized intervention and feedback requirements can significantly facilitate the development of robotic assistance systems in endovascular interventions. 
The modular design allows for the incorporation of extensions and additional features, potentially introducing new functionalities such as manipulation of the position and orientation of the fluoroscopy system to optimize device visibility or device force measurement by analyzing the contacts of device and vessel wall to provide force feedback. 
\textit{stEVE} can be adapted to support other endoluminal interventions by implementing the required organ lumen and device and adapting the simulation script to the new setup. 
For instance, with the SOFA simulation engine it is possible to include magnetic continuum robots \cite{dreyfus_simulation_2022} or cable driven devices \cite{renda_discrete_2018}. 

Furthermore, complex but interesting future developments include tissue deformation, tissue inhomogeneities, and motions due to heartbeat or breathing. 
Integrating tissue inhomogeneities can be achieved by adding additional rigid calcifications to the lumen of the vessel tree. 
Vessel deformation is highly relevant for interventions, such as endovascular aortic repair \cite{koutouzi_iliac_2019}. 
Considering deformation during interventions can improve radiation exposure, reduce the need for contrast agents, and shorten procedure time \cite{martel_move_2020}. 
This could be implemented by including a deformation model for the vessel mesh, allowing for vessel flexibility, albeit with increased computational demands. 
Heartbeat effects could be simulated by applying a force field on the device that exerts force in the direction of blood flow, while breathing motion could be simulated by applying cyclic movement and deformation patterns to the vessel system.

\section{Conclusion}
This work introduces the simulation framework \textit{stEVE}, tailored for endovascular interventions with the primary purpose of streamlining and supporting future research endeavors in the field of robotic assistance systems in endovascular interventions.

The introduced benchmarks are designed to enhance the comparability of research in the domain of learning-based autonomous endovascular navigation. They serve as a standardized experimental setup, fostering a common ground for researchers to evaluate and advance their approaches.

The experimental outcomes demonstrate the feasibility of \textit{stEVE} and underscore its potential for transferring controllers trained in simulation to real-world scenarios. 
Nevertheless, they also reveal areas that offer opportunities for further enhancement and exploration in future research endeavors, including the concurrent manipulation of two concentric devices and improvements in adaption to patient individual vessel geometries.

\footnotesize
\section*{Acknowledgments}
Harry Robertshaw and Benjamin Jackson were funded by King’s College London Centre for Doctoral Training on Surgical \& Interventional Engineering.

S.M.Hadi Sadati, Thomas Booth and Christos Bergeles were supported by an NIHR Cardiovascular MIC Grant awarded to King’s College London, and by core funding from the Wellcome/EPSRC Centre for Medical Engineering [WT203148/Z/16/Z].

\footnotesize
\section*{CRediT authorship contribution statement}
\textbf{Lennart Karstensen:} Conceptualization, Data curation, Formal analysis, Investigation, Methodology, Software, Visualization, Writing – original draft. 
\textbf{Harry Robertshaw:} Data curation, Investigation, Validation, Writing – review \& editing.
\textbf{Johannes~Hatzl:} Validation, Writing – review \& editing.
\textbf{Benjamin~Jackson:} Data curation, Validation, Writing – review \& editing.
\textbf{Jens~Langejürgen:} Project administration, Resources, Writing – review \& editing.
\textbf{Katharina~Breininger:} Supervision, Writing – review \& editing.
\textbf{Christian~Uhl:} Validation, Writing – review \& editing.
\textbf{S.M.Hadi~Sadati:} Conceptualization, Validation, Writing – review \& editing.
\textbf{Thomas~Booth:} Conceptualization, Funding acquisition, Supervision, Project administration, Writing – review \& editing.
\textbf{Christos~Bergeles:} Conceptualization, Funding acquisition, Writing – review \& editing.
\textbf{Franziska~Mathis-Ullrich:} Conceptualization, Project administration, Resources, Supervision, Writing – review \& editing.

\footnotesize
\section*{Declaration of competing interest}
The authors declare that they have no known competing financial
interests or personal relationships that could have appeared to influence
the work reported in this paper.

\footnotesize
\section*{Data availability}
Simulation framework, benchmarks and training scripts are available on GitHub: \href{https://github.com/lkarstensen}{https://github.com/lkarstensen}

\footnotesize
\section*{Declaration of Generative AI and AI-assisted technologies in the writing process}
During the preparation of this work the authors used Copilot (Microsoft, Redmond, Washington, United States), ChatGPT (OpenAI, San Francisco, United States) and DeepL Write (DeepL SE, Cologne, Germany) in order to improve grammar and readability of this paper. After using this tool/service, the authors reviewed and edited the content as needed and take full responsibility for the content of the publication.


\normalsize
\bibliography{main}

\begin{thebibliography}{50}
\providecommand{\natexlab}[1]{#1}
\providecommand{\url}[1]{\texttt{#1}}
\expandafter\ifx\csname urlstyle\endcsname\relax
  \providecommand{\doi}[1]{doi: #1}\else
  \providecommand{\doi}{doi: \begingroup \urlstyle{rm}\Url}\fi

\bibitem[Ho et~al.(2007)Ho, Cheng, Wu, Ting, Poon, Cheng, Mok, and
  Tsang]{ho_ionizing_2007}
Pei Ho, Stephen~W.K. Cheng, P.M. Wu, Albert~C.W. Ting, Jensen~T.C. Poon,
  Clement~K.M. Cheng, Joseph~H.M. Mok, and M.S. Tsang.
\newblock Ionizing radiation absorption of vascular surgeons during
  endovascular procedures.
\newblock \emph{Journal of Vascular Surgery}, 46\penalty0 (3):\penalty0
  455--459, September 2007.
\newblock ISSN 07415214.
\newblock \doi{10.1016/j.jvs.2007.04.034}.

\bibitem[Böckler(2020)]{bockler_praktische_2020}
D.~Böckler.
\newblock Praktische {Tipps} für den persönlichen {Strahlenschutz} bei
  endovaskulären {Eingriffen} im {Hybrid}-{Operationssaal}.
\newblock \emph{Gefässchirurgie}, 25\penalty0 (1):\penalty0 19--30, February
  2020.
\newblock ISSN 1434-3932.
\newblock \doi{10.1007/s00772-020-00620-9}.

\bibitem[Crinnion et~al.(2022)Crinnion, Jackson, Sood, Lynch, Bergeles, Liu,
  Rhode, Mendes~Pereira, and Booth]{crinnion_robotics_2022}
William Crinnion, Ben Jackson, Avnish Sood, Jeremy Lynch, Christos Bergeles,
  Hongbin Liu, Kawal Rhode, Vitor Mendes~Pereira, and Thomas~C. Booth.
\newblock Robotics in neurointerventional surgery: a systematic review of the
  literature.
\newblock \emph{Journal of neurointerventional surgery}, 14\penalty0
  (6):\penalty0 539--545, June 2022.
\newblock ISSN 1759-8486 1759-8478.
\newblock \doi{10.1136/neurintsurg-2021-018096}.
\newblock Place: England.

\bibitem[Goldstein et~al.(2004)Goldstein, Balter, Cowley, Hodgson, and
  Klein]{goldstein_occupational_2004}
James~A. Goldstein, Stephen Balter, Michael Cowley, John Hodgson, and Lloyd~W.
  Klein.
\newblock Occupational hazards of interventional cardiologists: {Prevalence} of
  orthopedic health problems in contemporary practice.
\newblock \emph{Catheterization and Cardiovascular Interventions}, 63\penalty0
  (4):\penalty0 407--411, December 2004.
\newblock ISSN 1522-1946, 1522-726X.
\newblock \doi{10.1002/ccd.20201}.
\newblock URL \url{https://onlinelibrary.wiley.com/doi/10.1002/ccd.20201}.

\bibitem[Yan et~al.(2022)Yan, Hu, Alcock, Ghrooda, Trivedi, McEachern,
  Kaderali, and Shankar]{yan_access_2022}
Yi~Yan, Kai Hu, Susan Alcock, Esseddeeg Ghrooda, Anurag Trivedi, James
  McEachern, Zul Kaderali, and Jai Shankar.
\newblock Access to {Endovascular} {Thrombectomy} for {Stroke} in {Rural}
  {Versus} {Urban} {Regions}.
\newblock \emph{Canadian Journal of Neurological Sciences}, 49\penalty0
  (1):\penalty0 70--75, 2022.
\newblock \doi{10.1017/cjn.2021.35}.
\newblock Publisher: Cambridge University Press.

\bibitem[Zhao et~al.(2022)Zhao, Mei, Luo, Mao, Zhao, Liu, and
  Wu]{zhao_remote_2022}
Yang Zhao, Ziyang Mei, Xiaoxiao Luo, Jingsong Mao, Qingliang Zhao, Gang Liu,
  and Dezhi Wu.
\newblock Remote vascular interventional surgery robotics: a literature review.
\newblock \emph{Quantitative Imaging in Medicine and Surgery}, 12\penalty0
  (4):\penalty0 2552--2574, April 2022.
\newblock ISSN 22234292, 22234306.
\newblock \doi{10.21037/qims-21-792}.

\bibitem[Mitros et~al.(2022)Mitros, Sadati, Henry, Da~Cruz, and
  Bergeles]{mitros_theoretical_2022}
Zisos Mitros, S.M.~Hadi Sadati, Ross Henry, Lyndon Da~Cruz, and Christos
  Bergeles.
\newblock From {Theoretical} {Work} to {Clinical} {Translation}: {Progress} in
  {Concentric} {Tube} {Robots}.
\newblock \emph{Annual Review of Control, Robotics, and Autonomous Systems},
  5\penalty0 (1):\penalty0 335--359, May 2022.
\newblock ISSN 2573-5144, 2573-5144.
\newblock \doi{10.1146/annurev-control-042920-014147}.
\newblock URL
  \url{https://www.annualreviews.org/doi/10.1146/annurev-control-042920-014147}.

\bibitem[Patel et~al.(2020)Patel, Shah, Soni, Radadiya, Patel, Tiwari, and
  Pancholy]{patel_comparison_2020}
Tejas~M. Patel, Sanjay~C. Shah, Yash~Y. Soni, Rajni~C. Radadiya, Gaurav~A.
  Patel, Pradyot~O. Tiwari, and Samir~B. Pancholy.
\newblock Comparison of {Robotic} {Percutaneous} {Coronary} {Intervention}
  {With} {Traditional} {Percutaneous} {Coronary} {Intervention}: {A}
  {Propensity} {Score}–{Matched} {Analysis} of a {Large} {Cohort}.
\newblock \emph{Circulation: Cardiovascular Interventions}, 13\penalty0
  (5):\penalty0 e008888, May 2020.
\newblock ISSN 1941-7640, 1941-7632.
\newblock \doi{10.1161/CIRCINTERVENTIONS.119.008888}.

\bibitem[Patel et~al.(2019)Patel, Shah, and Pancholy]{patel_long_2019}
Tejas~M. Patel, Sanjay~C. Shah, and Samir~B. Pancholy.
\newblock Long {Distance} {Tele}-{Robotic}-{Assisted} {Percutaneous} {Coronary}
  {Intervention}: {A} {Report} of {First}-in-{Human} {Experience}.
\newblock \emph{EClinicalMedicine}, 14:\penalty0 53--58, September 2019.
\newblock ISSN 25895370.
\newblock \doi{10.1016/j.eclinm.2019.07.017}.
\newblock URL
  \url{https://linkinghub.elsevier.com/retrieve/pii/S2589537019301373}.

\bibitem[Duan et~al.(2023)Duan, Akinyemi, Du, Ma, Chen, Wang, Omisore, Luo,
  Wang, and Wang]{duan_technical_2023}
Wenke Duan, Toluwanimi Akinyemi, Wenjing Du, Jun Ma, Xingyu Chen, Fuhao Wang,
  Olatunji Omisore, Jingjing Luo, Hongbo Wang, and Lei Wang.
\newblock Technical and {Clinical} {Progress} on {Robot}-{Assisted}
  {Endovascular} {Interventions}: {A} {Review}.
\newblock \emph{Micromachines}, 14\penalty0 (1):\penalty0 197, January 2023.
\newblock ISSN 2072-666X.
\newblock \doi{10.3390/mi14010197}.
\newblock URL \url{https://www.mdpi.com/2072-666X/14/1/197}.

\bibitem[Attanasio et~al.(2021)Attanasio, Scaglioni, De~Momi, Fiorini, and
  Valdastri]{attanasio_autonomy_2021}
Aleks Attanasio, Bruno Scaglioni, Elena De~Momi, Paolo Fiorini, and Pietro
  Valdastri.
\newblock Autonomy in {Surgical} {Robotics}.
\newblock \emph{Annual Review of Control, Robotics, and Autonomous Systems},
  4\penalty0 (1):\penalty0 651--679, May 2021.
\newblock ISSN 2573-5144, 2573-5144.
\newblock \doi{10.1146/annurev-control-062420-090543}.
\newblock URL
  \url{https://www.annualreviews.org/doi/10.1146/annurev-control-062420-090543}.

\bibitem[Pore et~al.(2023)Pore, Li, Dall'Alba, Hernansanz, De~Momi, Menciassi,
  Casals~Gelpí, Dankelman, Fiorini, and Poorten]{pore_autonomous_2023}
Ameya Pore, Zhen Li, Diego Dall'Alba, Albert Hernansanz, Elena De~Momi, Arianna
  Menciassi, Alicia Casals~Gelpí, Jenny Dankelman, Paolo Fiorini, and
  Emmanuel~Vander Poorten.
\newblock Autonomous {Navigation} for {Robot}-{Assisted} {Intraluminal} and
  {Endovascular} {Procedures}: {A} {Systematic} {Review}.
\newblock \emph{IEEE Transactions on Robotics}, 39\penalty0 (4):\penalty0
  2529--2548, August 2023.
\newblock ISSN 1552-3098, 1941-0468.
\newblock \doi{10.1109/TRO.2023.3269384}.

\bibitem[Ji et~al.(2011)Ji, Hou, and Xie]{ji_guidewire_2011}
Cheng Ji, Zeng-Guang Hou, and Xiao-Liang Xie.
\newblock Guidewire navigation and delivery system for robot-assisted
  cardiology interventions.
\newblock In \emph{{IEEE} 10th {International} {Conference} on {Cognitive}
  {Informatics} and {Cognitive} {Computing} ({ICCI}-{CC}'11)}, pages 330--335,
  Banff, AB, Canada, August 2011. IEEE.
\newblock ISBN 978-1-4577-1695-9.
\newblock \doi{10.1109/COGINF.2011.6016161}.
\newblock URL \url{http://ieeexplore.ieee.org/document/6016161/}.

\bibitem[Jayender et~al.(2008)Jayender, Azizian, and
  Patel]{jayender_autonomous_2008}
J.~Jayender, M.~Azizian, and R.V. Patel.
\newblock Autonomous {Image}-{Guided} {Robot}-{Assisted} {Active} {Catheter}
  {Insertion}.
\newblock \emph{IEEE Transactions on Robotics}, 24\penalty0 (4):\penalty0
  858--871, August 2008.
\newblock ISSN 1552-3098.
\newblock \doi{10.1109/TRO.2008.2001353}.
\newblock URL \url{http://ieeexplore.ieee.org/document/4598902/}.

\bibitem[Li et~al.(2022)Li, Feridooni, Cuen-Ojeda, Kishibe, de~Mestral,
  Mamdani, and Al-Omran]{li_machine_2022}
Ben Li, Tiam Feridooni, Cesar Cuen-Ojeda, Teruko Kishibe, Charles de~Mestral,
  Muhammad Mamdani, and Mohammed Al-Omran.
\newblock Machine learning in vascular surgery: a systematic review and
  critical appraisal.
\newblock \emph{npj Digital Medicine}, 5\penalty0 (1):\penalty0 1--10, January
  2022.
\newblock ISSN 2398-6352.
\newblock \doi{10.1038/s41746-021-00552-y}.
\newblock URL \url{https://www.nature.com/articles/s41746-021-00552-y}.

\bibitem[Ritter et~al.(2022)Ritter, Karstensen, Langejürgen, Hatzl,
  Mathis-Ullrich, and Uhl]{ritter_quality-dependent_2022}
Jacqueline Ritter, Lennart Karstensen, Jens Langejürgen, Johannes Hatzl,
  Franziska Mathis-Ullrich, and Christian Uhl.
\newblock Quality-dependent {Deep} {Learning} for {Safe} {Autonomous}
  {Guidewire} {Navigation}.
\newblock \emph{Current Directions in Biomedical Engineering}, 8\penalty0
  (1):\penalty0 21--24, July 2022.
\newblock ISSN 2364-5504.
\newblock \doi{10.1515/cdbme-2022-0006}.

\bibitem[Meng et~al.(2022)Meng, Guo, Zhou, and Chen]{meng_evaluation_2022}
Fanxu Meng, Shuxiang Guo, Wei Zhou, and Zhengyang Chen.
\newblock Evaluation of an {Autonomous} {Navigation} {Method} for {Vascular}
  {Interventional} {Surgery} in {Virtual} {Environment}.
\newblock In \emph{2022 {IEEE} {International} {Conference} on {Mechatronics}
  and {Automation} ({ICMA})}, pages 1599--1604, 2022.
\newblock \doi{10.1109/ICMA54519.2022.9856107}.

\bibitem[Schegg et~al.(2022)Schegg, Dequidt, Coevoet, Leurent, Sabatier, Preux,
  and Duriez]{schegg_automated_2022}
Pierre Schegg, Jeremie Dequidt, Eulalie Coevoet, Edouard Leurent, Remi
  Sabatier, Philippe Preux, and Christian Duriez.
\newblock Automated {Planning} for {Robotic} {Guidewire} {Navigation} in the
  {Coronary} {Arteries}.
\newblock In \emph{2022 {IEEE} 5th {International} {Conference} on {Soft}
  {Robotics} ({RoboSoft})}, pages 239--246, Edinburgh, United Kingdom, April
  2022. IEEE.
\newblock ISBN 978-1-66540-828-8.
\newblock \doi{10.1109/RoboSoft54090.2022.9762096}.
\newblock URL \url{https://ieeexplore.ieee.org/document/9762096/}.

\bibitem[Karstensen et~al.(2023)Karstensen, Ritter, Hatzl, Ernst, Langejürgen,
  Uhl, and Mathis-Ullrich]{karstensen_recurrent_2023}
Lennart Karstensen, Jacqueline Ritter, Johannes Hatzl, Floris Ernst, Jens
  Langejürgen, Christian Uhl, and Franziska Mathis-Ullrich.
\newblock Recurrent neural networks for generalization towards the vessel
  geometry in autonomous endovascular guidewire navigation in the aortic arch.
\newblock \emph{International Journal of Computer Assisted Radiology and
  Surgery}, May 2023.
\newblock ISSN 1861-6429.
\newblock \doi{10.1007/s11548-023-02938-7}.

\bibitem[Jianu et~al.(2024)Jianu, Huang, Vu, Abdelaziz, Fichera, Lee,
  Berthet-Rayne, Baena, and Nguyen]{jianu_cathsim_2024}
Tudor Jianu, Baoru Huang, Minh~Nhat Vu, Mohamed E. M.~K. Abdelaziz, Sebastiano
  Fichera, Chun-Yi Lee, Pierre Berthet-Rayne, Ferdinando Rodriguez~Y Baena, and
  Anh Nguyen.
\newblock {CathSim}: {An} {Open}-{Source} {Simulator} for {Endovascular}
  {Intervention}.
\newblock \emph{IEEE Transactions on Medical Robotics and Bionics}, pages 1--1,
  2024.
\newblock ISSN 2576-3202.
\newblock \doi{10.1109/TMRB.2024.3421256}.
\newblock URL \url{https://ieeexplore.ieee.org/document/10587130/}.

\bibitem[Kienzlen et~al.(2022)Kienzlen, Jaensch, Verl, and
  Cheng]{kienzlen_concept_2022}
Annika Kienzlen, Florian Jaensch, Alexander Verl, and Leo Cheng.
\newblock Concept for a {Reinforcement} {Learning} {Approach} to {Navigate}
  {Catheters} {Through} {Blood} {Vessels}.
\newblock In \emph{2022 28th {International} {Conference} on {Mechatronics} and
  {Machine} {Vision} in {Practice} ({M2VIP})}, pages 1--4, 2022.
\newblock \doi{10.1109/M2VIP55626.2022.10041096}.

\bibitem[Scarponi et~al.(2024)Scarponi, Duprez, Nageotte, and
  Cotin]{scarponi_zero-shot_2024}
Valentina Scarponi, Michel Duprez, Florent Nageotte, and Stéphane Cotin.
\newblock A zero-shot reinforcement learning strategy for autonomous guidewire
  navigation.
\newblock \emph{International Journal of Computer Assisted Radiology and
  Surgery}, April 2024.
\newblock ISSN 1861-6429.
\newblock \doi{10.1007/s11548-024-03092-4}.
\newblock URL \url{https://link.springer.com/10.1007/s11548-024-03092-4}.

\bibitem[Zhao et~al.(2019)Zhao, Guo, Wang, Cui, Ma, Zeng, Liu, Jiang, Li, Shi,
  and Xiao]{zhao_cnn-based_2019}
Yan Zhao, Shuxiang Guo, Yuxin Wang, Jinxin Cui, Youchun Ma, Yuwen Zeng, Xinke
  Liu, Yuhua Jiang, Youxinag Li, Liwei Shi, and Nan Xiao.
\newblock A {CNN}-based prototype method of unstructured surgical state
  perception and navigation for an endovascular surgery robot.
\newblock \emph{Medical \& Biological Engineering \& Computing}, 57\penalty0
  (9):\penalty0 1875--1887, September 2019.
\newblock ISSN 1741-0444.
\newblock \doi{10.1007/s11517-019-02002-0}.

\bibitem[Kweon et~al.(2021)Kweon, Kim, Lee, Kwon, Park, Song, Kim, Park, Back,
  Roh, Moon, Choi, and Kim]{kweon_deep_2021}
Jihoon Kweon, Kyunghwan Kim, Chaehyuk Lee, Hwi Kwon, Jinwoo Park, Kyoseok Song,
  Young~In Kim, Jeeone Park, Inwook Back, Jae-Hyung Roh, Youngjin Moon, Jaesoon
  Choi, and Young-Hak Kim.
\newblock Deep {Reinforcement} {Learning} for {Guidewire} {Navigation} in
  {Coronary} {Artery} {Phantom}.
\newblock \emph{IEEE Access}, 9:\penalty0 166409--166422, 2021.
\newblock \doi{10.1109/ACCESS.2021.3135277}.

\bibitem[Li et~al.(2023)Li, Zhou, Xie, Liu, Gui, Xiang, Wang, and
  Hou]{li_discrete_2023}
Hao Li, Xiao-Hu Zhou, Xiao-Liang Xie, Shi-Qi Liu, Mei-Jiang Gui, Tian-Yu Xiang,
  Jin-Li Wang, and Zeng-Guang Hou.
\newblock Discrete soft actor-critic with auto-encoder on vascular robotic
  system.
\newblock \emph{Robotica}, 41\penalty0 (4):\penalty0 1115--1126, 2023.
\newblock \doi{10.1017/S0263574722001527}.
\newblock Publisher: Cambridge University Press.

\bibitem[Li et~al.(2024)Li, Zhou, Xie, Liu, Feng, and Hou]{li_casog_2024}
Hao Li, Xiao-Hu Zhou, Xiao-Liang Xie, Shi-Qi Liu, Zhen-Qiu Feng, and Zeng-Guang
  Hou.
\newblock {CASOG}: {Conservative} {Actor}–{Critic} {With} {SmOoth} {Gradient}
  for {Skill} {Learning} in {Robot}-{Assisted} {Intervention}.
\newblock \emph{IEEE Transactions on Industrial Electronics}, 71\penalty0
  (7):\penalty0 7725--7734, July 2024.
\newblock ISSN 0278-0046, 1557-9948.
\newblock \doi{10.1109/TIE.2023.3310021}.
\newblock URL \url{https://ieeexplore.ieee.org/document/10254299/}.

\bibitem[Wang et~al.(2022)Wang, Liu, Shu, Cao, Zhang, and Xie]{wang_study_2022}
Shuang Wang, Zheng Liu, Xiongpeng Shu, Yongfeng Cao, Ling Zhang, and Le~Xie.
\newblock Study on {Autonomous} {Delivery} of {Guidewire} {Based} on {Improved}
  {YOLOV5s} on {Vascular} {Model} {Platform}.
\newblock In \emph{2022 {IEEE} {International} {Conference} on {Robotics} and
  {Biomimetics} ({ROBIO})}, pages 1--6, 2022.
\newblock \doi{10.1109/ROBIO55434.2022.10011829}.

\bibitem[Yang et~al.(2022)Yang, Song, and Hu]{yang_guidewire_2022}
Deai Yang, Jingzhou Song, and Yuhang Hu.
\newblock Guidewire feeding method based on deep reinforcement learning for
  vascular intervention robot.
\newblock In \emph{2022 {IEEE} {International} {Conference} on {Mechatronics}
  and {Automation} ({ICMA})}, pages 1287--1293, 2022.
\newblock \doi{10.1109/ICMA54519.2022.9856351}.

\bibitem[Song et~al.(2022)Song, Yi, Won, and Woo]{song_learning-based_2022}
Hwa-Seob Song, Byung-Ju Yi, Jong~Yun Won, and Jaehong Woo.
\newblock Learning-based catheter and guidewire-driven autonomous vascular
  intervention robotic system for reduced repulsive force.
\newblock \emph{Journal of Computational Design and Engineering}, 9\penalty0
  (5):\penalty0 1549--1564, October 2022.
\newblock ISSN 2288-5048.
\newblock \doi{10.1093/jcde/qwac074}.

\bibitem[Chi et~al.(2020)Chi, Dagnino, Kwok, Nguyen, Kundrat, Abdelaziz, Riga,
  Bicknell, and Yang]{chi_collaborative_2020}
Wenqiang Chi, Giulio Dagnino, Trevor M.~Y. Kwok, Anh Nguyen, Dennis Kundrat,
  Mohamed E. M.~K. Abdelaziz, Celia Riga, Colin Bicknell, and Guang-Zhong Yang.
\newblock Collaborative {Robot}-{Assisted} {Endovascular} {Catheterization}
  with {Generative} {Adversarial} {Imitation} {Learning}.
\newblock In \emph{2020 {IEEE} {International} {Conference} on {Robotics} and
  {Automation} ({ICRA})}, pages 2414--2420, 2020.
\newblock \doi{10.1109/ICRA40945.2020.9196912}.

\bibitem[Karstensen et~al.(2022)Karstensen, Ritter, Hatzl, Pätz, Langejürgen,
  Uhl, and Mathis-Ullrich]{karstensen_learning-based_2022}
Lennart Karstensen, Jacqueline Ritter, Johannes Hatzl, Torben Pätz, Jens
  Langejürgen, Christian Uhl, and Franziska Mathis-Ullrich.
\newblock Learning-based autonomous vascular guidewire navigation without human
  demonstration in the venous system of a porcine liver.
\newblock \emph{International Journal of Computer Assisted Radiology and
  Surgery}, May 2022.
\newblock ISSN 1861-6429.
\newblock \doi{10.1007/s11548-022-02646-8}.

\bibitem[Robertshaw et~al.(2023)Robertshaw, Karstensen, Jackson, Sadati, Rhode,
  Ourselin, Granados, and Booth]{robertshaw_artificial_2023}
Harry Robertshaw, Lennart Karstensen, Benjamin Jackson, Hadi Sadati, Kawal
  Rhode, Sebastien Ourselin, Alejandro Granados, and Thomas~C. Booth.
\newblock Artificial intelligence in the autonomous navigation of endovascular
  interventions: a systematic review.
\newblock \emph{Frontiers in Human Neuroscience}, 17:\penalty0 1239374, August
  2023.
\newblock ISSN 1662-5161.
\newblock \doi{10.3389/fnhum.2023.1239374}.

\bibitem[Hong et~al.(2020)Hong, Kao, Kuo, Wang, Chang, and
  Shih]{hong_cholecseg8k_2020}
W.-Y. Hong, C.-L. Kao, Y.-H. Kuo, J.-R. Wang, W.-L. Chang, and C.-S. Shih.
\newblock {CholecSeg8k}: {A} {Semantic} {Segmentation} {Dataset} for
  {Laparoscopic} {Cholecystectomy} {Based} on {Cholec80}, December 2020.
\newblock URL \url{http://arxiv.org/abs/2012.12453}.
\newblock arXiv:2012.12453 [cs].

\bibitem[Towers et~al.(2023)Towers, Terry, Kwiatkowski, Balis, Cola, Deleu,
  Goulão, Kallinteris, KG, Krimmel, Perez-Vicente, Pierré, Schulhoff, Tai,
  Tan, and Younis]{towers_gymnasium_2023}
Mark Towers, Jordan~K Terry, Ariel Kwiatkowski, John~U. Balis, Gianluca Cola,
  Tristan Deleu, Manuel Goulão, Andreas Kallinteris, Arjun KG, Markus Krimmel,
  Rodrigo Perez-Vicente, Andrea Pierré, Sander Schulhoff, Jun~Jet Tai, Andrew
  Jin~Shen Tan, and Omar~G. Younis.
\newblock Gymnasium, August 2023.
\newblock URL \url{https://doi.org/10.5281/zenodo.8269265}.

\bibitem[Molinero et~al.(2019)Molinero, Dagnino, Liu, Chi, Abdelaziz, Kwok,
  Riga, and Yang]{molinero_haptic_2019}
M.~B. Molinero, G.~Dagnino, J.~Liu, W.~Chi, M.~E. M.~K. Abdelaziz, T.M.Y. Kwok,
  C.~Riga, and G.Z. Yang.
\newblock Haptic {Guidance} for {Robot}-{Assisted} {Endovascular} {Procedures}:
  {Implementation} and {Evaluation} on {Surgical} {Simulator}.
\newblock In \emph{2019 {IEEE}/{RSJ} {International} {Conference} on
  {Intelligent} {Robots} and {Systems} ({IROS})}, pages 5398--5403, Macau,
  China, November 2019. IEEE.
\newblock ISBN 978-1-72814-004-9.
\newblock \doi{10.1109/IROS40897.2019.8967712}.
\newblock URL \url{https://ieeexplore.ieee.org/document/8967712/}.

\bibitem[Brilakis(2021{\natexlab{a}})]{brilakis_chapter_2021}
Emmanouil Brilakis.
\newblock Chapter 8 - {Wiring}.
\newblock In Emmanouil Brilakis, editor, \emph{Manual of {Percutaneous}
  {Coronary} {Interventions}}, pages 123--139. Academic Press,
  2021{\natexlab{a}}.
\newblock ISBN 978-0-12-819367-9.
\newblock \doi{https://doi.org/10.1016/B978-0-12-819367-9.00008-1}.

\bibitem[Brilakis(2021{\natexlab{b}})]{brilakis_chapter_2021-1}
Emmanouil Brilakis.
\newblock Chapter 30 - {Equipment}.
\newblock In Emmanouil Brilakis, editor, \emph{Manual of {Percutaneous}
  {Coronary} {Interventions}}, pages 487--574. Academic Press,
  2021{\natexlab{b}}.
\newblock ISBN 978-0-12-819367-9.
\newblock \doi{https://doi.org/10.1016/B978-0-12-819367-9.00030-5}.

\bibitem[Schneider(2019)]{schneider_endovascular_2019}
Peter Schneider.
\newblock \emph{Endovascular skills: guidewire and catheter skills for
  endovascular surgery}.
\newblock CRC press, Boca Raton, 4 edition, 2019.
\newblock ISBN 978-0-429-15630-4.

\bibitem[Wilson et~al.(2013)Wilson, Ortiz, and Johnson]{wilson_vascular_2013}
Nathan~M. Wilson, Ana~K. Ortiz, and Allison~B. Johnson.
\newblock The {Vascular} {Model} {Repository}: {A} {Public} {Resource} of
  {Medical} {Imaging} {Data} and {Blood} {Flow} {Simulation} {Results}.
\newblock \emph{Journal of Medical Devices}, 7\penalty0 (040923), December
  2013.
\newblock ISSN 1932-6181.
\newblock \doi{10.1115/1.4025983}.

\bibitem[Baba(2023)]{baba_yahya_simmons_2023}
Yahya Baba.
\newblock Simmons catheter, December 2023.
\newblock URL \url{https://radiopaedia.org/articles/simmons-catheter}.

\bibitem[Cowen(2018)]{cowen_cardiovascular_2018}
Arnold~R. Cowen.
\newblock Cardiovascular {X}-ray {Imaging}: {Physics}, {Equipment}, and
  {Techniques}.
\newblock In Peter Lanzer, editor, \emph{Textbook of {Catheter}-{Based}
  {Cardiovascular} {Interventions}: {A} {Knowledge}-{Based} {Approach}}, pages
  147--198. Springer International Publishing, Cham, 2 edition, 2018.
\newblock ISBN 978-3-319-55994-0.
\newblock \doi{10.1007/978-3-319-55994-0_11}.

\bibitem[Faure et~al.(2012)Faure, Duriez, Delingette, Allard, Gilles,
  Marchesseau, Talbot, Courtecuisse, Bousquet, Peterlik, and
  Cotin]{faure_sofa_2012}
François Faure, Christian Duriez, Hervé Delingette, Jérémie Allard,
  Benjamin Gilles, Stéphanie Marchesseau, Hugo Talbot, Hadrien Courtecuisse,
  Guillaume Bousquet, Igor Peterlik, and Stéphane Cotin.
\newblock {SOFA}: {A} {Multi}-{Model} {Framework} for {Interactive} {Physical}
  {Simulation}.
\newblock In Yohan Payan, editor, \emph{Soft {Tissue} {Biomechanical}
  {Modeling} for {Computer} {Assisted} {Surgery}}, pages 283--321. Springer
  Berlin Heidelberg, Berlin, Heidelberg, 2012.
\newblock ISBN 978-3-642-29014-5.
\newblock \doi{10.1007/8415_2012_125}.

\bibitem[Wei et~al.(2012)Wei, Cotin, Dequidt, Duriez, Allard, and
  Kerrien]{wei_near_2012}
Yiyi Wei, Stéphane Cotin, Jérémie Dequidt, Christian Duriez, Jérémie
  Allard, and Erwan Kerrien.
\newblock A ({Near}) {Real}-{Time} {Simulation} {Method} of {Aneurysm} {Coil}
  {Embolization}.
\newblock In Yasuo Murai, editor, \emph{Aneurysm}, pages 223--248. InTech,
  August 2012.
\newblock \doi{10.5772/48635}.

\bibitem[Abdelaal et~al.(2014)Abdelaal, Plourde, MacHaalany, Arsenault, Rimac,
  Déry, Barbeau, Larose, De~Larochellière, Nguyen, Allende, Ribeiro,
  Costerousse, Mongrain, and Bertrand]{abdelaal_effectiveness_2014}
Eltigani Abdelaal, Guillaume Plourde, Jimmy MacHaalany, Jean Arsenault, Goran
  Rimac, Jean-Pierre Déry, Gérald Barbeau, Eric Larose, Robert
  De~Larochellière, Can~M. Nguyen, Ricardo Allende, Henrique Ribeiro, Olivier
  Costerousse, Rosaire Mongrain, and Olivier~F. Bertrand.
\newblock Effectiveness of {Low} {Rate} {Fluoroscopy} at {Reducing} {Operator}
  and {Patient} {Radiation} {Dose} {During} {Transradial} {Coronary}
  {Angiography} and {Interventions}.
\newblock \emph{JACC: Cardiovascular Interventions}, 7\penalty0 (5):\penalty0
  567--574, May 2014.
\newblock ISSN 19368798.
\newblock \doi{10.1016/j.jcin.2014.02.005}.

\bibitem[Haarnoja et~al.(2018)Haarnoja, Zhou, Hartikainen, Tucker, Ha, Tan,
  Kumar, Zhu, Gupta, Abbeel, and Levine]{haarnoja_soft_2018}
Tuomas Haarnoja, Aurick Zhou, Kristian Hartikainen, George Tucker, Sehoon Ha,
  Jie Tan, Vikash Kumar, Henry Zhu, Abhishek Gupta, Pieter Abbeel, and Sergey
  Levine.
\newblock Soft {Actor}-{Critic} {Algorithms} and {Applications}.
\newblock \emph{arXiv e-prints}, page arXiv:1812.05905, December 2018.
\newblock \doi{10.48550/arXiv.1812.05905}.
\newblock \_eprint: 1812.05905.

\bibitem[Eyberg et~al.(2022)Eyberg, Karstensen, Pusch, Horsch, and
  Langejürgen]{eyberg_ros2-based_2022}
Christoph Eyberg, Lennart Karstensen, Tim Pusch, Johannes Horsch, and Jens
  Langejürgen.
\newblock A {ROS2}-based {Testbed} {Environment} for {Endovascular} {Robotic}
  {Systems}.
\newblock \emph{Current Directions in Biomedical Engineering}, 8\penalty0
  (1):\penalty0 89--92, July 2022.
\newblock ISSN 2364-5504.
\newblock \doi{10.1515/cdbme-2022-0023}.

\bibitem[Dreyfus et~al.(2022)Dreyfus, Boehler, and
  Nelson]{dreyfus_simulation_2022}
Roland Dreyfus, Quentin Boehler, and Bradley~J. Nelson.
\newblock A {Simulation} {Framework} for {Magnetic} {Continuum} {Robots}.
\newblock \emph{IEEE Robotics and Automation Letters}, 7\penalty0 (3):\penalty0
  8370--8376, July 2022.
\newblock ISSN 2377-3766, 2377-3774.
\newblock \doi{10.1109/LRA.2022.3187249}.
\newblock URL \url{https://ieeexplore.ieee.org/document/9810340/}.

\bibitem[Renda et~al.(2018)Renda, Boyer, Dias, and
  Seneviratne]{renda_discrete_2018}
Federico Renda, Frederic Boyer, Jorge Dias, and Lakmal Seneviratne.
\newblock Discrete {Cosserat} {Approach} for {Multisection} {Soft}
  {Manipulator} {Dynamics}.
\newblock \emph{IEEE Transactions on Robotics}, 34\penalty0 (6):\penalty0
  1518--1533, December 2018.
\newblock ISSN 1552-3098, 1941-0468.
\newblock \doi{10.1109/TRO.2018.2868815}.
\newblock URL \url{https://ieeexplore.ieee.org/document/8500341/}.

\bibitem[Koutouzi et~al.(2019)Koutouzi, Pfister, Breininger, Hellström, Roos,
  and Falkenberg]{koutouzi_iliac_2019}
Giasemi Koutouzi, Marcus Pfister, Katharina Breininger, Mikael Hellström,
  Håkan Roos, and Mårten Falkenberg.
\newblock Iliac artery deformation during {EVAR}.
\newblock \emph{Vascular}, 27\penalty0 (5):\penalty0 511--517, October 2019.
\newblock ISSN 1708-5381, 1708-539X.
\newblock \doi{10.1177/1708538119840565}.
\newblock URL \url{http://journals.sagepub.com/doi/10.1177/1708538119840565}.

\bibitem[Breininger et~al.(2020)Breininger, Pfister, Kowarschik, and
  Maier]{martel_move_2020}
Katharina Breininger, Marcus Pfister, Markus Kowarschik, and Andreas Maier.
\newblock Move {Over} {There}: {One}-{Click} {Deformation} {Correction} for
  {Image} {Fusion} {During} {Endovascular} {Aortic} {Repair}.
\newblock In Anne~L. Martel, Purang Abolmaesumi, Danail Stoyanov, Diana Mateus,
  Maria~A. Zuluaga, S.~Kevin Zhou, Daniel Racoceanu, and Leo Joskowicz,
  editors, \emph{Medical {Image} {Computing} and {Computer} {Assisted}
  {Intervention} – {MICCAI} 2020}, volume 12264, pages 713--723, Cham, 2020.
  Springer International Publishing.
\newblock ISBN 978-3-030-59718-4 978-3-030-59719-1.
\newblock \doi{10.1007/978-3-030-59719-1_69}.
\newblock URL \url{https://link.springer.com/10.1007/978-3-030-59719-1_69}.
\newblock Series Title: Lecture Notes in Computer Science.

\end{thebibliography}


\end{document}